\documentclass[11pt, a4paper, logo, copyright, nonumbering]{deepseek}
\usepackage[authoryear, sort&compress, round]{natbib}
\usepackage{dblfloatfix}
\usepackage{ulem}
\usepackage{caption}
\usepackage{dramatist}
\usepackage{xspace}
\usepackage{pifont} 
\usepackage{multirow}
\usepackage{tcolorbox}
\usepackage{xltabular}
\usepackage{longtable}
\usepackage{hyperref}
\usepackage{amsfonts}
\usepackage{amsmath}
\usepackage{amssymb}
\usepackage{lineno}
\usepackage{multirow}
\usepackage{adjustbox}

\usepackage[bottom]{footmisc}

\usepackage{CJKutf8}
\usepackage{subfigure}
\usepackage{setspace}

\usepackage{dsfont}
\usepackage{array} 
\usepackage{tabularx} 
\usepackage{subfigure} 
\usepackage{xcolor} 

\usepackage{lipsum}  
\usepackage{multicol} 

\makeatletter
\def\@BTrule[#1]{%
  \ifx\longtable\undefined
    \let\@BTswitch\@BTnormal
  \else\ifx\hline\LT@hline
    \nobreak
    \let\@BTswitch\@BLTrule
  \else
     \let\@BTswitch\@BTnormal
  \fi\fi
  \global\@thisrulewidth=#1\relax
  \ifnum\@thisruleclass=\tw@\vskip\@aboverulesep\else
  \ifnum\@lastruleclass=\z@\vskip\@aboverulesep\else
  \ifnum\@lastruleclass=\@ne\vskip\doublerulesep\fi\fi\fi
  \@BTswitch}
\makeatother

\addto\extrasenglish{
}

 {\begin{list}{}%
         {\setlength{\leftmargin}{#1}}%
         \item[]%
 }
 {\end{list}}
 
\reportnumber{001} 

\renewcommand{\today}{}

\newcommand{\dsvii}{DeepSeek-V2}

\newcommand{\dsattn}{MLA}
\newcommand{\dsmoe}{DeepSeekMoE}

\newcommand{\dsviii}{DeepSeek-V3}

\title{\centering \dsviii{} Technical Report}

\author[*]{
DeepSeek-AI
\\
\small
\texttt{research@deepseek.com}
}

\renewcommand{\phi}{\varphi}

\renewcommand{\leq}{\leqslant}

\renewcommand{\epsilon}{\varepsilon}
\renewcommand{\imath}{\mathrm{i}}

\newlength{\restsubwidth}
\newlength{\restsubheight}
\newlength{\restsubmoreheight}
\newcommand{\rest}[2]{%
        \settowidth{\restsubwidth}{\ensuremath{#2}}
        \settoheight{\restsubheight}{\ensuremath{{}_{#2}}}
        \ensuremath{{#1\hskip 0.5pt}_{\vrule\kern2pt\parbox[b][%
        4pt][b]{\the\restsubwidth}{%
                        \ensuremath{{}_{#2}}}}}
        }

\begin{abstract}

We present \dsviii{}, a strong Mixture-of-Experts~(MoE) language model with 671B total parameters with 37B activated for each token. 
To achieve efficient inference and cost-effective training, \dsviii{} adopts Multi-head Latent Attention~(MLA) and \dsmoe{} architectures, which were thoroughly validated in \dsvii{}. 
Furthermore, \dsviii{} pioneers an auxiliary-loss-free strategy for load balancing and sets a multi-token prediction training objective for stronger performance. 
We pre-train \dsviii{} on 14.8 trillion diverse and high-quality tokens, followed by Supervised Fine-Tuning and Reinforcement Learning stages to fully harness its capabilities. 
Comprehensive evaluations reveal that \dsviii{} outperforms other open-source models and achieves performance comparable to leading closed-source models.
Despite its excellent performance, \dsviii{} requires only 2.788M H800 GPU hours for its full training.
In addition, its training process is remarkably stable. 
Throughout the entire training process, we did not experience any irrecoverable loss spikes or perform any rollbacks. 
The model checkpoints are available at \url{https://github.com/deepseek-ai/DeepSeek-V3}. 

\end{abstract}

\begin{document}
\begin{CJK*}{UTF8}{gbsn}

\maketitle

\begin{figure}[h]
\centering
\includegraphics[width=\textwidth]{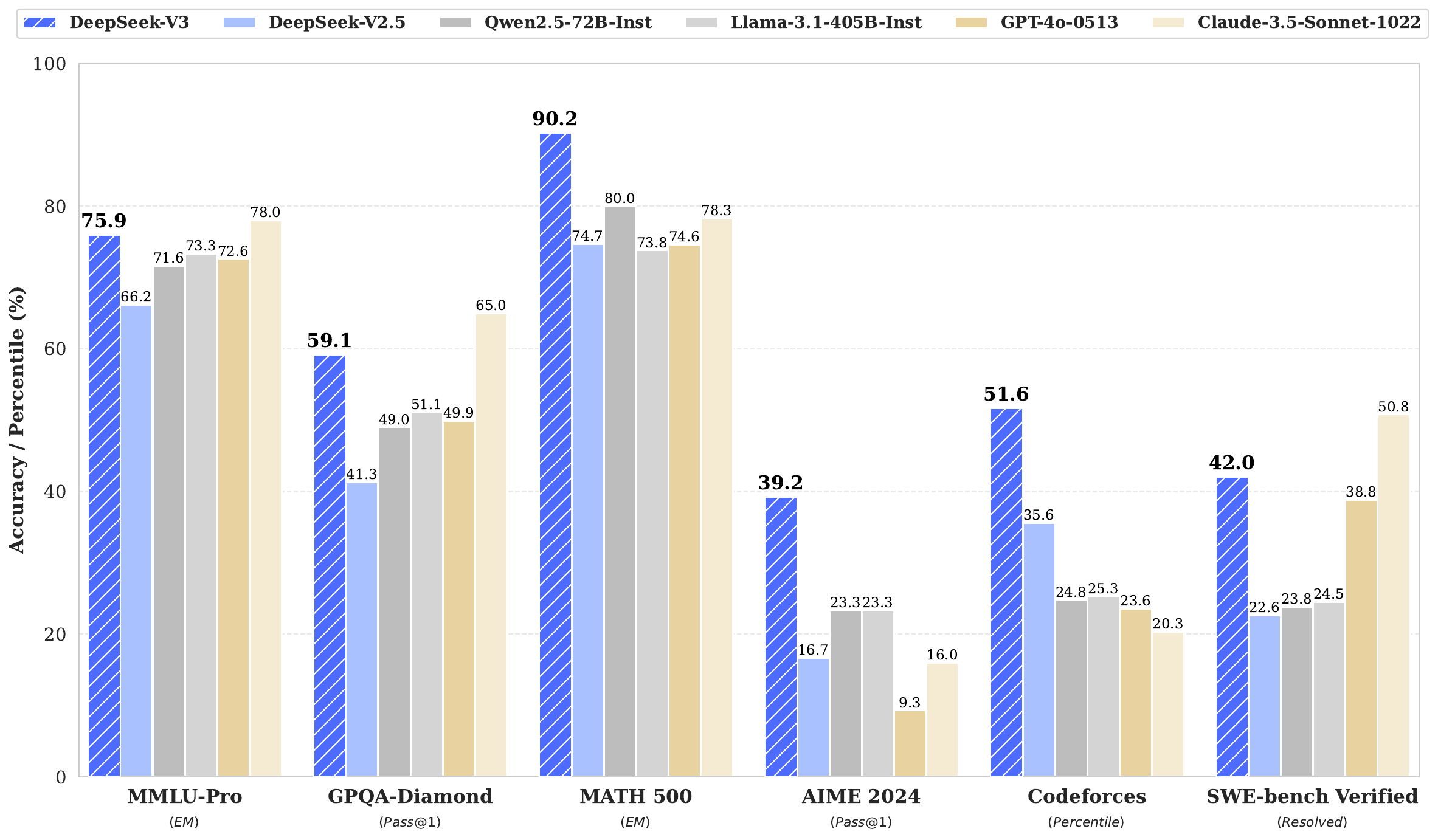}
\caption{
    \centering
    Benchmark performance of \dsviii{} and its counterparts.
}
\label{fig:dsv3_performance}
\end{figure}

\newpage

\begin{spacing}{0.9}
\tableofcontents
\end{spacing}

\newpage

\section{Introduction}

In recent years, Large Language Models~(LLMs) have been undergoing rapid iteration and evolution~\citep{gpt4o,claude35sonnet,gemini1_5}, progressively diminishing the gap towards Artificial General Intelligence~(AGI).
Beyond closed-source models, open-source models, including DeepSeek series~\citep{dsvi,dsvii,dscodervi,dscodervii}, LLaMA series~\citep{llama,llama2,llama3,llama3_1_405b}, Qwen series~\citep{qwen,qwen1_5,qwen2_5}, and Mistral series~\citep{mistral,mixtral8x22b}, are also making significant strides, endeavoring to close the gap with their closed-source counterparts.
To further push the boundaries of open-source model capabilities, we scale up our models and introduce \dsviii{}, a large Mixture-of-Experts~(MoE) model with 671B parameters, of which 37B are activated for each token.

With a forward-looking perspective, we consistently strive for strong model performance and economical costs.
Therefore, in terms of architecture, \dsviii{} still adopts Multi-head Latent Attention~(\dsattn{})~\citep{dsvii} for efficient inference and \dsmoe{}~\citep{deepseekmoe} for cost-effective training. 
These two architectures have been validated in \dsvii{}~\citep{dsvii}, demonstrating their capability to maintain robust model performance while achieving efficient training and inference.
Beyond the basic architecture, we implement two additional strategies to further enhance the model capabilities. 
Firstly, \dsviii{} pioneers an auxiliary-loss-free strategy~\citep{noaux_tc} for load balancing, with the aim of minimizing the adverse impact on model performance that arises from the effort to encourage load balancing.
Secondly, \dsviii{} employs a multi-token prediction training objective, which we have observed to enhance the overall performance on evaluation benchmarks.

In order to achieve efficient training, we support the FP8 mixed precision training and implement comprehensive optimizations for the training framework.
Low-precision training has emerged as a promising solution for efficient training~\citep{bf16train, fp16train, fp8lm, llm.int8}, its evolution being closely tied to advancements in hardware capabilities~\citep{fp8format, hifp8format, microscaling}. 
In this work, we introduce an FP8 mixed precision training framework and, for the first time, validate its effectiveness on an extremely large-scale model. 
Through the support for FP8 computation and storage, we achieve both accelerated training and reduced GPU memory usage. 
As for the training framework, we design the DualPipe algorithm for efficient pipeline parallelism, which has fewer pipeline bubbles and hides most of the communication during training through computation-communication overlap.
This overlap ensures that, as the model further scales up, as long as we maintain a constant computation-to-communication ratio, we can still employ fine-grained experts across nodes while achieving a near-zero all-to-all communication overhead.
In addition, we also develop efficient cross-node all-to-all communication kernels to fully utilize InfiniBand~(IB) and NVLink bandwidths. 
Furthermore, we meticulously optimize the memory footprint, making it possible to train \dsviii{} without using costly tensor parallelism.
Combining these efforts, we achieve high training efficiency. 

During pre-training, we train \dsviii{} on 14.8T high-quality and diverse tokens. 
The pre-training process is remarkably stable. 
Throughout the entire training process, we did not encounter any irrecoverable loss spikes or have to roll back.
Next, we conduct a two-stage context length extension for \dsviii{}. In the first stage, the maximum context length is extended to 32K, and in the second stage, it is further extended to 128K. 
Following this, we conduct post-training, including Supervised Fine-Tuning~(SFT) and Reinforcement Learning~(RL) on the base model of \dsviii{}, to align it with human preferences and further unlock its potential. 
During the post-training stage, we distill the reasoning capability from the DeepSeek-R1 series of models, and meanwhile carefully maintain the balance between model accuracy and generation length. 

We evaluate \dsviii{} on a comprehensive array of benchmarks. 
Despite its economical training costs, comprehensive evaluations reveal that \dsviii{}-Base has emerged as the strongest open-source base model currently available, especially in code and math. 
Its chat version also outperforms other open-source models and achieves performance comparable to leading closed-source models, including GPT-4o and Claude-3.5-Sonnet, on a series of standard and open-ended benchmarks. 

\begin{table}[t]
    \centering
    \setlength{\tabcolsep}{6pt}
    \begin{tabular}{l | c c c | c}
        \toprule
        \textbf{Training Costs} & \textbf{Pre-Training} & \textbf{Context Extension} & \textbf{Post-Training} & \textbf{Total} \\
        \midrule
        in H800 GPU Hours & 2664K & 119K & 5K & 2788K \\
        in USD & \$5.328M & \$0.238M & \$0.01M & \$5.576M \\
        \bottomrule
    \end{tabular}
    \caption{
    Training costs of \dsviii{}, assuming the rental price of H800 is \$2 per GPU hour.
    }
    \label{tab:cost}
\end{table}

Lastly, we emphasize again the economical training costs of \dsviii{}, summarized in Table~\ref{tab:cost}, achieved through our optimized co-design of algorithms, frameworks, and hardware. 
During the pre-training stage, training \dsviii{} on each trillion tokens requires only 180K H800 GPU hours, i.e., 3.7 days on our cluster with 2048 H800 GPUs. 
Consequently, our pre-training stage is completed in less than two months and costs 2664K GPU hours.
Combined with 119K GPU hours for the context length extension and 5K GPU hours for post-training, \dsviii{} costs only 2.788M GPU hours for its full training. 
Assuming the rental price of the H800 GPU is \$2 per GPU hour, our total training costs amount to only \$5.576M.
Note that the aforementioned costs include only the official training of \dsviii{}, excluding the costs associated with prior research and ablation experiments on architectures, algorithms, or data. 

Our main contribution includes:

\noindent
\textbf{Architecture: Innovative Load Balancing Strategy and Training Objective}
\begin{itemize}[topsep=0pt]
    \item 
    On top of the efficient architecture of \dsvii{}, we pioneer an auxiliary-loss-free strategy for load balancing, which minimizes the performance degradation that arises from encouraging load balancing.
    \item 
    We investigate a Multi-Token Prediction (MTP) objective and prove it beneficial to model performance. 
    It can also be used for speculative decoding for inference acceleration. 
\end{itemize}

\noindent
\textbf{Pre-Training: Towards Ultimate Training Efficiency}
\begin{itemize}[topsep=0pt]
    \item
    We design an FP8 mixed precision training framework and, for the first time, validate the feasibility and effectiveness of FP8 training on an extremely large-scale model. 
    \item 
    Through the co-design of algorithms, frameworks, and hardware, we overcome the communication bottleneck in cross-node MoE training, achieving near-full computation-communication overlap. 
    This significantly enhances our training efficiency and reduces the training costs, enabling us to further scale up the model size without additional overhead.
    \item 
    At an economical cost of only 2.664M H800 GPU hours, we complete the pre-training of \dsviii{} on 14.8T tokens, producing the currently strongest open-source base model. 
    The subsequent training stages after pre-training require only 0.1M GPU hours.
\end{itemize}

\noindent
\textbf{Post-Training: Knowledge Distillation from DeepSeek-R1}
\begin{itemize}[topsep=0pt]
    \item 
    We introduce an innovative methodology to distill reasoning capabilities from the long-Chain-of-Thought (CoT) model, specifically from one of the DeepSeek R1 series models, into standard LLMs, particularly \dsviii{}.
    Our pipeline elegantly incorporates the verification and reflection patterns of R1 into \dsviii{} and notably improves its reasoning performance.
    Meanwhile, we also maintain control over the output style and length of \dsviii{}. 
\end{itemize}
\noindent
\textbf{Summary of Core Evaluation Results}
\begin{itemize}[topsep=0pt]
    \item \textbf{Knowledge}:
    (1) 
    On educational benchmarks such as MMLU, MMLU-Pro, and GPQA, \dsviii{} outperforms all other open-source models, achieving 88.5 on MMLU, 75.9 on MMLU-Pro, and 59.1 on GPQA.
    Its performance is comparable to leading closed-source models like GPT-4o and Claude-Sonnet-3.5, narrowing the gap between open-source and closed-source models in this domain.
    (2)
    For factuality benchmarks, \dsviii{} demonstrates superior performance among open-source models on both SimpleQA and Chinese SimpleQA. 
    While it trails behind GPT-4o and Claude-Sonnet-3.5 in English factual knowledge (SimpleQA), it surpasses these models in Chinese factual knowledge (Chinese SimpleQA), highlighting its strength in Chinese factual knowledge.
    
    \item \textbf{Code, Math, and Reasoning}: 
    (1)
    \dsviii{} achieves state-of-the-art performance on math-related benchmarks among all non-long-CoT open-source and closed-source models. 
    Notably, it even outperforms o1-preview on specific benchmarks, such as MATH-500, demonstrating its robust mathematical reasoning capabilities. 
    (2)
    On coding-related tasks, \dsviii{} emerges as the top-performing model for coding competition benchmarks, such as LiveCodeBench, solidifying its position as the leading model in this domain. 
    For engineering-related tasks, while \dsviii{} performs slightly below Claude-Sonnet-3.5, it still outpaces all other models by a significant margin, demonstrating its competitiveness across diverse technical benchmarks.
\end{itemize}

In the remainder of this paper, we first present a detailed exposition of our \dsviii{} model architecture (Section~\ref{sec:arch}). 
Subsequently, we introduce our infrastructures, encompassing our compute clusters, the training framework, the support for FP8 training, the inference deployment strategy, and our suggestions on future hardware design.
Next, we describe our pre-training process, including the construction of training data, hyper-parameter settings, long-context extension techniques, the associated evaluations, as well as some discussions (Section~\ref{sec:pre-training}). 
Thereafter, we discuss our efforts on post-training, which include Supervised Fine-Tuning (SFT), Reinforcement Learning (RL), the corresponding evaluations, and discussions (Section~\ref{sec:alignment}). 
Lastly, we conclude this work, discuss existing limitations of \dsviii{}, and propose potential directions for future research (Section~\ref{sec:conclusion}).

\section{Architecture}
\label{sec:arch}

We first introduce the basic architecture of \dsviii{}, featured by Multi-head Latent Attention~(\dsattn{})~\citep{dsvii} for efficient inference and \dsmoe{}~\citep{deepseekmoe} for economical training.
Then, we present a Multi-Token Prediction (MTP) training objective, which we have observed to enhance the overall performance on evaluation benchmarks.
For other minor details not explicitly mentioned, \dsviii{} adheres to the settings of \dsvii{}~\citep{dsvii}. 

\begin{figure}[!t]
\centering
\includegraphics[width=0.99\linewidth]{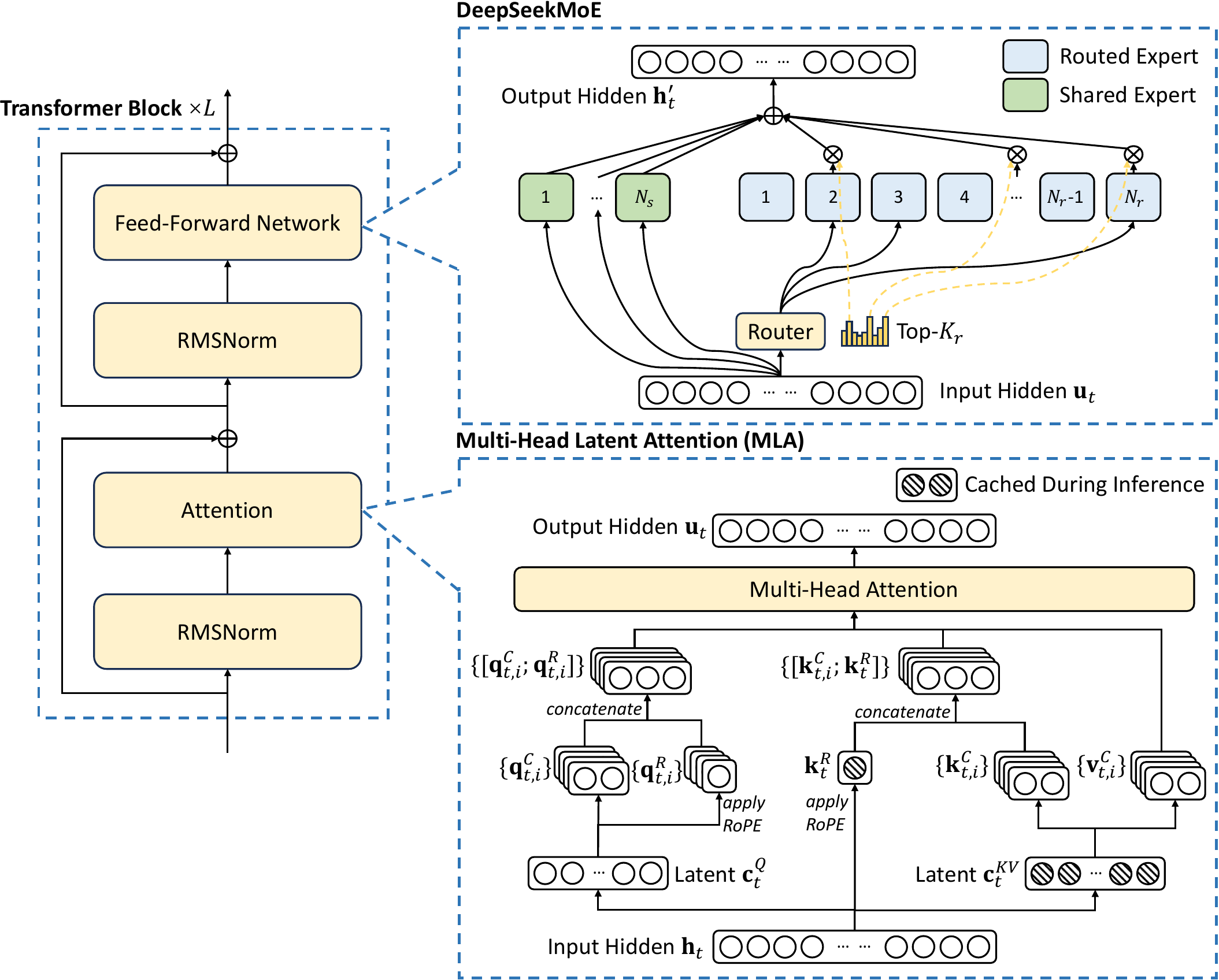}
\caption{
    Illustration of the basic architecture of \dsviii{}. 
    Following \dsvii{}, we adopt \dsattn{} and \dsmoe{} for efficient inference and economical training.
}
\label{fig:basic_arch}
\end{figure}

\subsection{Basic Architecture}

The basic architecture of \dsviii{} is still within the Transformer~\citep{transformer} framework. 
For efficient inference and economical training, \dsviii{} also adopts \dsattn{} and \dsmoe{}, which have been thoroughly validated by \dsvii{}. 
Compared with \dsvii{}, an exception is that we additionally introduce an auxiliary-loss-free load balancing strategy~\citep{noaux_tc} for \dsmoe{} to mitigate the performance degradation induced by the effort to ensure load balance. 
Figure~\ref{fig:basic_arch} illustrates the basic architecture of \dsviii{}, and we will briefly review the details of MLA and DeepSeekMoE in this section. 

\subsubsection{Multi-Head Latent Attention}

For attention, \dsviii{} adopts the \dsattn{} architecture. 
Let $d$ denote the embedding dimension, $n_h$ denote the number of attention heads, $d_h$ denote the dimension per head, and $\mathbf{h}_{t} \in \mathbb{R}^{d}$ denote the attention input for the $t$-th token at a given attention layer.
The core of \dsattn{} is the low-rank joint compression for attention keys and values to reduce Key-Value (KV) cache during inference:
\begin{align}
    \boxed{\color{blue} \mathbf{c}_{t}^{KV}} &= W^{DKV} \mathbf{h}_{t}, \\
    [\mathbf{k}_{t, 1}^{C};\mathbf{k}_{t, 2}^{C};...;\mathbf{k}_{t, n_{h}}^{C}] = \mathbf{k}_{t}^{C} &= W^{UK} \mathbf{c}_{t}^{KV}, \\
    \boxed{\color{blue}\mathbf{k}_{t}^{R}} &= \operatorname{RoPE}({W^{KR}} \mathbf{h}_{t}), \\
    \mathbf{k}_{t, i} &= [\mathbf{k}_{t, i}^{C}; \mathbf{k}_{t}^{R}], \\
    [\mathbf{v}_{t, 1}^{C};\mathbf{v}_{t, 2}^{C};...;\mathbf{v}_{t, n_{h}}^{C}] = \mathbf{v}_{t}^{C} &= W^{UV} \mathbf{c}_{t}^{KV}, 
\end{align}
where $\mathbf{c}_{t}^{KV} \in \mathbb{R}^{d_c}$ is the compressed latent vector for keys and values; 
$d_c (\ll d_h n_h)$ indicates the KV compression dimension;
$W^{DKV} \in \mathbb{R}^{d_c \times d}$ denotes the down-projection matrix;
$W^{UK},W^{UV} \in \mathbb{R}^{d_h n_h \times d_c}$ are the up-projection matrices for keys and values, respectively;
$W^{KR} \in \mathbb{R}^{d_h^R \times d}$ is the matrix used to produce the decoupled key that carries Rotary Positional Embedding (RoPE)~\citep{su2024roformer}; 
$\operatorname{RoPE}(\cdot)$ denotes the operation that applies RoPE matrices; 
and $[\cdot;\cdot]$ denotes concatenation.
Note that for MLA, only the blue-boxed vectors (i.e., $\color{blue} \mathbf{c}_{t}^{KV}$ and $\color{blue}\mathbf{k}_{t}^{R}$) need to be cached during generation, which results in significantly reduced KV cache while maintaining performance comparable to standard Multi-Head Attention (MHA)~\citep{transformer}.

For the attention queries, we also perform a low-rank compression, which can reduce the activation memory during training:
\begin{align}
    \mathbf{c}_{t}^{Q} &= W^{DQ} \mathbf{h}_{t}, \\
    [\mathbf{q}_{t, 1}^{C};\mathbf{q}_{t, 2}^{C};...;\mathbf{q}_{t, n_{h}}^{C}] = \mathbf{q}_{t}^{C} &= W^{UQ} \mathbf{c}_{t}^{Q}, \\
    [\mathbf{q}_{t, 1}^{R};\mathbf{q}_{t, 2}^{R};...;\mathbf{q}_{t, n_{h}}^{R}] = \mathbf{q}_{t}^{R} &= \operatorname{RoPE}({W^{QR}} \mathbf{c}_{t}^{Q}), \\
    \mathbf{q}_{t, i} &= [\mathbf{q}_{t, i}^{C}; \mathbf{q}_{t, i}^{R}],
\end{align}
where $\mathbf{c}_{t}^{Q} \in \mathbb{R}^{d_c^{\prime}}$ is the compressed latent vector for queries; 
$d_c^{\prime} (\ll d_h n_h)$ denotes the query compression dimension; 
$W^{DQ} \in \mathbb{R}^{d_c^{\prime} \times d}, W^{UQ} \in \mathbb{R}^{d_h n_h \times d_c^{\prime}}$ are the down-projection and up-projection matrices for queries, respectively;
and $W^{QR} \in \mathbb{R}^{d_h^R n_h \times d_c^{\prime}}$ is the matrix to produce the decoupled queries that carry RoPE. 

Ultimately, the attention queries ($\mathbf{q}_{t, i}$), keys ($\mathbf{k}_{j, i}$), and values ($\mathbf{v}_{j, i}^{C}$) are combined to yield the final attention output $\mathbf{u}_{t}$:
\begin{align}
    \mathbf{o}_{t, i} &= \sum_{j=1}^{t} \operatorname{Softmax}_j(\frac{\mathbf{q}_{t, i}^T \mathbf{k}_{j, i}}{\sqrt{d_{h} + d_{h}^{R}}}) \mathbf{v}_{j, i}^{C}, \\
    \mathbf{u}_{t} &= W^{O} [\mathbf{o}_{t, 1};\mathbf{o}_{t, 2};...;\mathbf{o}_{t, n_{h}}],
\end{align}
where $W^{O} \in \mathbb{R}^{d \times d_h n_h}$ denotes the output projection matrix. 

\subsubsection{\dsmoe{} with Auxiliary-Loss-Free Load Balancing}

\paragraph{Basic Architecture of \dsmoe{}.}
For Feed-Forward Networks~(FFNs), \dsviii{} employs the \dsmoe{} architecture~\citep{deepseekmoe}. 
Compared with traditional MoE architectures like GShard~\citep{gshard}, \dsmoe{} uses finer-grained experts and isolates some experts as shared ones.
Let $\mathbf{u}_{t}$ denote the FFN input of the $t$-th token, we compute the FFN output $\mathbf{h}_{t}^{\prime}$ as follows:
\begin{align}
    \mathbf{h}_{t}^{\prime} & = \mathbf{u}_{t} + \sum_{i=1}^{N_{s}} {\operatorname{FFN}^{(s)}_{i}\left( \mathbf{u}_{t} \right)} + \sum_{i=1}^{N_r} {g_{i,t} \operatorname{FFN}^{(r)}_{i}\left( \mathbf{u}_{t} \right)}, \\
    g_{i,t} & = \frac{g^{\prime}_{i,t}}{\sum_{j=1}^{N_r} g^{\prime}_{j,t}}, \\
    g^{\prime}_{i,t} & = \begin{cases} 
    s_{i,t}, & s_{i,t} \in \operatorname{Topk} (\{ s_{j, t} | 1 \leq j \leq N_r \}, K_{r}), \\
    0, & \text{otherwise}, 
    \end{cases} \\
    s_{i,t} & = \operatorname{Sigmoid} \left( {\mathbf{u}_{t}}^{T} \mathbf{e}_{i} \right),
\end{align}
where $N_{s}$ and $N_r$ denote the numbers of shared experts and routed experts, respectively; 
$\operatorname{FFN}^{(s)}_{i}(\cdot)$ and $\operatorname{FFN}^{(r)}_{i}(\cdot)$ denote the $i$-th shared expert and the $i$-th routed expert, respectively; 
$K_{r}$ denotes the number of activated routed experts; 
$g_{i,t}$ is the gating value for the $i$-th expert; 
$s_{i,t}$ is the token-to-expert affinity; 
$\mathbf{e}_{i}$ is the centroid vector of the $i$-th routed expert; 
and $\operatorname{Topk}(\cdot, K)$ denotes the set comprising $K$ highest scores among the affinity scores calculated for the $t$-th token and all routed experts.
Slightly different from \dsvii{}, \dsviii{} uses the sigmoid function to compute the affinity scores, and applies a normalization among all selected affinity scores to produce the gating values. 

\paragraph{Auxiliary-Loss-Free Load Balancing.}
For MoE models, an unbalanced expert load will lead to routing collapse~\citep{moe} and diminish computational efficiency in scenarios with expert parallelism. 
Conventional solutions usually rely on the auxiliary loss~\citep{switch,gshard} to avoid unbalanced load. 
However, too large an auxiliary loss will impair the model performance~\citep{noaux_tc}. 
To achieve a better trade-off between load balance and model performance, we pioneer an auxiliary-loss-free load balancing strategy~\citep{noaux_tc} to ensure load balance. 
To be specific, we introduce a bias term $b_i$ for each expert and add it to the corresponding affinity scores $s_{i,t}$ to determine the top-K routing:
\begin{align}
    g^{\prime}_{i,t} & = \begin{cases} 
    s_{i,t}, & s_{i,t} + b_i \in \operatorname{Topk} (\{ s_{j, t} + b_j | 1 \leq j \leq N_r \}, K_{r}), \\
    0, & \text{otherwise}.
    \end{cases}
\end{align}
Note that the bias term is only used for routing.
The gating value, which will be multiplied with the FFN output, is still derived from the original affinity score $s_{i,t}$.
During training, we keep monitoring the expert load on the whole batch of each training step.
At the end of each step, we will decrease the bias term by $\gamma$ if its corresponding expert is overloaded, and increase it by $\gamma$ if its corresponding expert is underloaded, where $\gamma$ is a hyper-parameter called bias update speed.
Through the dynamic adjustment, \dsviii{} keeps balanced expert load during training, and achieves better performance than models that encourage load balance through pure auxiliary losses.

\paragraph{Complementary Sequence-Wise Auxiliary Loss.}
Although \dsviii{} mainly relies on the auxiliary-loss-free strategy for load balance, to prevent extreme imbalance within any single sequence, we also employ a complementary sequence-wise balance loss:
\begin{align}
    \mathcal{L}_{\mathrm{Bal}} & = \alpha \sum_{i=1}^{N_r}{f_i P_i}, \\
    f_i = \frac{N_r}{K_r T} \sum_{t=1}^{T} \mathds{1} & \left( s_{i,t} \in \operatorname{Topk} ( \{ s_{j, t} | 1 \leq j \leq N_r \}, K_{r} ) \right), \\
    s^{\prime}_{i,t} & = \frac{s_{i,t}}{\sum_{j=1}^{N_r} s_{j,t}}, \\
    P_i & = \frac{1}{T} \sum_{t=1}^{T}{s^{\prime}_{i,t}},
\end{align}
where the balance factor $\alpha$ is a hyper-parameter, which will be assigned an extremely small value for \dsviii{}; 
$\mathds{1}(\cdot)$ denotes the indicator function; 
and $T$ denotes the number of tokens in a sequence. 
The sequence-wise balance loss encourages the expert load on each sequence to be balanced. 

\paragraph{Node-Limited Routing.}
Like the device-limited routing used by \dsvii{}, \dsviii{} also uses a restricted routing mechanism to limit communication costs during training. 
In short, we ensure that each token will be sent to at most $M$ nodes, which are selected according to the sum of the highest $\frac{K_r}{M}$ affinity scores of the experts distributed on each node.
Under this constraint, our MoE training framework can nearly achieve full computation-communication overlap. 

\paragraph{No Token-Dropping.}
Due to the effective load balancing strategy, \dsviii{} keeps a good load balance during its full training. 
Therefore, \dsviii{} does not drop any tokens during training. 
In addition, we also implement specific deployment strategies to ensure inference load balance, so \dsviii{} also does not drop tokens during inference. 

\begin{figure}[!t]
\centering
\includegraphics[width=0.99\linewidth]{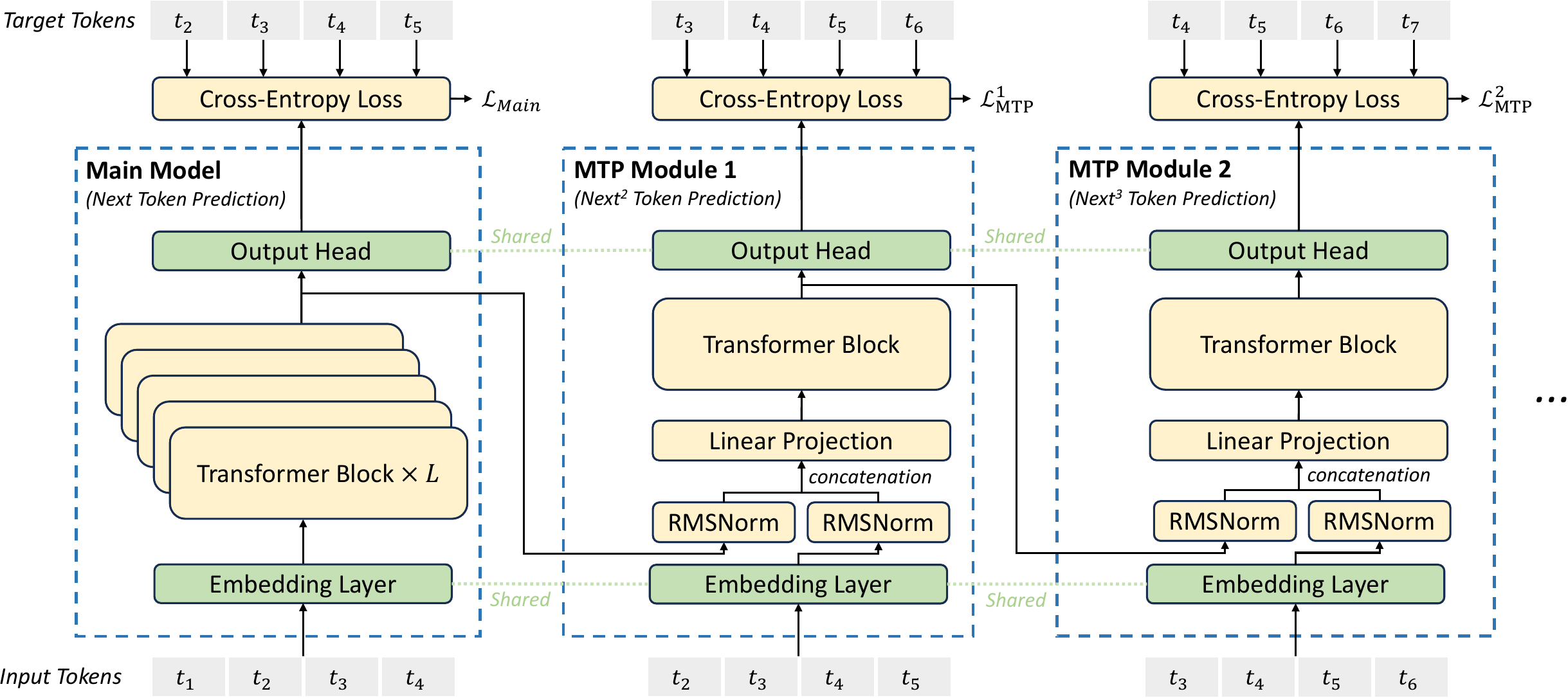}
\caption{
    Illustration of our Multi-Token Prediction (MTP) implementation. 
    We keep the complete causal chain for the prediction of each token at each depth. 
}
\label{fig:nextn}
\end{figure}

\subsection{Multi-Token Prediction}

Inspired by \citet{meta_mtp}, we investigate and set a Multi-Token Prediction (MTP) objective for \dsviii{}, which extends the prediction scope to multiple future tokens at each position.
On the one hand, an MTP objective densifies the training signals and may improve data efficiency.
On the other hand, MTP may enable the model to pre-plan its representations for better prediction of future tokens.
Figure~\ref{fig:nextn} illustrates our implementation of MTP.
Different from \citet{meta_mtp}, which parallelly predicts $D$ additional tokens using independent output heads, we sequentially predict additional tokens and keep the complete causal chain at each prediction depth.
We introduce the details of our MTP implementation in this section.

\paragraph{MTP Modules.}
To be specific, our MTP implementation uses $D$ sequential modules to predict $D$ additional tokens. 
The $k$-th MTP module consists of a shared embedding layer $\operatorname{Emb}(\cdot)$, a shared output head $\operatorname{OutHead}(\cdot)$, a Transformer block $\operatorname{TRM}_k(\cdot)$, and a projection matrix $M_k \in \mathbb{R}^{d \times 2d}$. 
For the $i$-th input token $t_i$, at the $k$-th prediction depth, we first combine the representation of the $i$-th token at the $(k-1)$-th depth $\mathbf{h}_i^{k-1} \in \mathbb{R}^{d}$ and the embedding of the $(i+k)$-th token $Emb(t_{i+k}) \in \mathbb{R}^{d}$ with the linear projection: 
\begin{equation}
    \mathbf{h}_i^{\prime k} = M_k [\operatorname{RMSNorm}(\mathbf{h}_i^{k-1}) ; \operatorname{RMSNorm}(\operatorname{Emb}(t_{i+k}))],
\end{equation}
where $[\cdot ; \cdot]$ denotes concatenation. 
Especially, when $k=1$, $\mathbf{h}_i^{k-1}$ refers to the representation given by the main model.
Note that for each MTP module, its embedding layer is shared with the main model. 
The combined $\mathbf{h}_i^{\prime k}$ serves as the input of the Transformer block at the $k$-th depth to produce the output representation at the current depth $\mathbf{h}_{i}^{k}$:
\begin{equation}
    \mathbf{h}_{1:T-k}^{k} = \operatorname{TRM}_k(\mathbf{h}_{1:T-k}^{\prime k}),
\end{equation}
where $T$ represents the input sequence length and $_{i:j}$ denotes the slicing operation (inclusive of both the left and right boundaries). 
Finally, taking $\mathbf{h}_{i}^{k}$ as the input, the shared output head will compute the probability distribution for the $k$-th additional prediction token $P_{i+1+k}^{k} \in \mathbb{R}^{V}$, where $V$ is the vocabulary size:
\begin{equation}
    P_{i+k+1}^{k} = \operatorname{OutHead}(\mathbf{h}_{i}^{k}).
\end{equation}
The output head $\operatorname{OutHead}(\cdot)$ linearly maps the representation to logits and subsequently applies the $\operatorname{Softmax}(\cdot)$ function to compute the prediction probabilities of the $k$-th additional token. 
Also, for each MTP module, its output head is shared with the main model. 
Our principle of maintaining the causal chain of predictions is similar to that of EAGLE~\citep{eagle}, but its primary objective is speculative decoding~\citep{speculative_xhm,speculative_google}, whereas we utilize MTP to improve training.

\paragraph{MTP Training Objective.}
For each prediction depth, we compute a cross-entropy loss $\mathcal{L}_{\text{MTP}}^{k}$:
\begin{equation}
    \mathcal{L}_{\text{MTP}}^{k} = \operatorname{CrossEntropy}(P_{2 + k:T + 1}^{k}, t_{2 + k:T + 1}) = -\frac{1}{T} \sum_{i=2 + k}^{T + 1} \log P_i^k [t_i],
\end{equation}
where $T$ denotes the input sequence length, $t_i$ denotes the ground-truth token at the $i$-th position, and $P_i^k [t_i]$ denotes the corresponding prediction probability of $t_i$, given by the $k$-th MTP module. 
Finally, we compute the average of the MTP losses across all depths and multiply it by a weighting factor $\lambda$ to obtain the overall MTP loss $\mathcal{L}_{\text{MTP}}$, which serves as an additional training objective for \dsviii{}:
\begin{equation}
    \mathcal{L}_{\text{MTP}} = \frac{\lambda}{D} \sum_{k=1}^{D} \mathcal{L}_{\text{MTP}}^{k}.
\end{equation}

\paragraph{MTP in Inference.}
Our MTP strategy mainly aims to improve the performance of the main model, so during inference, we can directly discard the MTP modules and the main model can function independently and normally.
Additionally, we can also repurpose these MTP modules for speculative decoding to further improve the generation latency.

\section{Infrastructures}
\label{sec:infra}

\subsection{Compute Clusters}

\dsviii{} is trained on a cluster equipped with 2048 NVIDIA H800 GPUs. 
Each node in the H800 cluster contains 8 GPUs connected by NVLink and NVSwitch within nodes. 
Across different nodes, InfiniBand~(IB) interconnects are utilized to facilitate communications.

\subsection{Training Framework}

The training of \dsviii{} is supported by the HAI-LLM framework, an efficient and lightweight training framework crafted by our engineers from the ground up. 
On the whole, \dsviii{} applies 16-way Pipeline Parallelism (PP)~\citep{qi2023zero}, 64-way Expert Parallelism (EP)~\citep{gshard} spanning 8 nodes, and ZeRO-1 Data Parallelism (DP)~\citep{zero}. 

In order to facilitate efficient training of \dsviii{}, we implement meticulous engineering optimizations.
Firstly, we design the DualPipe algorithm for efficient pipeline parallelism.
Compared with existing PP methods, DualPipe has fewer pipeline bubbles. 
More importantly, it overlaps the computation and communication phases across forward and backward processes, thereby addressing the challenge of heavy communication overhead introduced by cross-node expert parallelism. 
Secondly, we develop efficient cross-node all-to-all communication kernels to fully utilize IB and NVLink bandwidths and conserve Streaming Multiprocessors (SMs) dedicated to communication. 
Finally, we meticulously optimize the memory footprint during training, thereby enabling us to train \dsviii{} without using costly Tensor Parallelism~(TP).

\subsubsection{DualPipe and Computation-Communication Overlap}

\begin{figure}[t]
    \centering
    \includegraphics[width=0.9\textwidth]{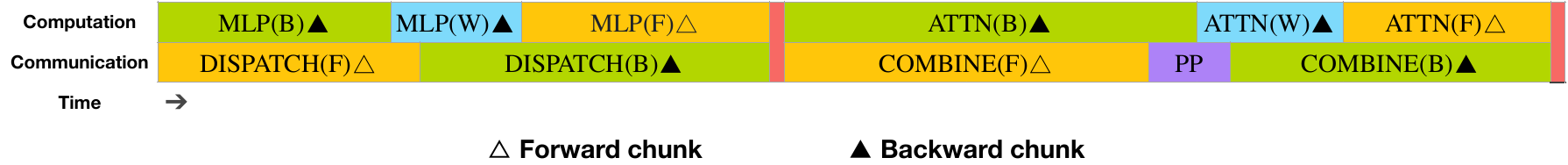}
    \caption{
        Overlapping strategy for a pair of individual forward and backward chunks (the boundaries of the transformer blocks are not aligned). 
        Orange denotes forward, green denotes "backward for input", blue denotes "backward for weights", purple denotes PP communication, and red denotes barriers.
        Both all-to-all and PP communication can be fully hidden. 
    }
    \label{fig:transformer-overlap}
\end{figure}

For \dsviii{}, the communication overhead introduced by cross-node expert parallelism results in an inefficient computation-to-communication ratio of approximately 1:1. 
To tackle this challenge, we design an innovative pipeline parallelism algorithm called DualPipe, which not only accelerates model training by effectively overlapping forward and backward computation-communication phases, but also reduces the pipeline bubbles.

\begin{figure}[t]
    \centering
    \includegraphics[width=0.99\linewidth]{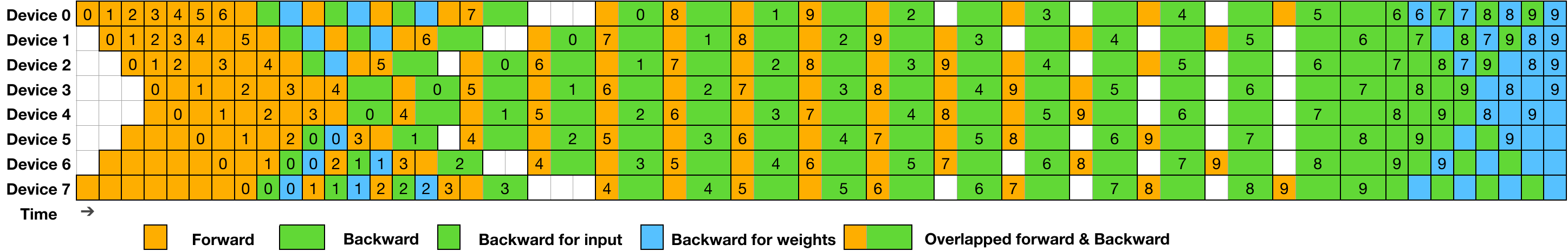}
    \caption{
        Example DualPipe scheduling for 8 PP ranks and 20 micro-batches in two directions. 
        The micro-batches in the reverse direction are symmetric to those in the forward direction, so we omit their batch ID for illustration simplicity. 
        Two cells enclosed by a shared black border have mutually overlapped computation and communication. 
    }
    \label{fig:dualpipe-schedules}
\end{figure}

The key idea of DualPipe is to overlap the computation and communication within a pair of individual forward and backward chunks.
To be specific, we divide each chunk into four components: \texttt{attention}, \texttt{all-to-all dispatch}, \texttt{MLP}, and \texttt{all-to-all combine}. 
Specially, for a backward chunk, both \texttt{attention} and \texttt{MLP} are further split into two parts, \texttt{backward for input} and \texttt{backward for weights}, like in ZeroBubble~\citep{zerobubble}. 
In addition, we have a \texttt{PP communication} component. 
As illustrated in Figure~\ref{fig:transformer-overlap}, for a pair of forward and backward chunks, we rearrange these components and manually adjust the ratio of GPU SMs dedicated to communication versus computation. 
In this overlapping strategy, we can ensure that both all-to-all and PP communication can be fully hidden during execution. 
Given the efficient overlapping strategy, the full DualPipe scheduling is illustrated in Figure~\ref{fig:dualpipe-schedules}. 
It employs a bidirectional pipeline scheduling, which feeds micro-batches from both ends of the pipeline simultaneously and a significant portion of communications can be fully overlapped.
This overlap also ensures that, as the model further scales up, as long as we maintain a constant computation-to-communication ratio, we can still employ fine-grained experts across nodes while achieving a near-zero all-to-all communication overhead.

In addition, even in more general scenarios without a heavy communication burden, DualPipe still exhibits efficiency advantages. 
In Table~\ref{tab:dualpipe-bubble}, we summarize the pipeline bubbles and memory usage across different PP methods. 
As shown in the table, compared with ZB1P~\citep{zerobubble} and 1F1B~\citep{pipedream}, DualPipe significantly reduces the pipeline bubbles while only increasing the peak activation memory by $\frac{1}{PP}$ times.
Although DualPipe requires keeping two copies of the model parameters, this does not significantly increase the memory consumption since we use a large EP size during training. 
Compared with Chimera~\citep{chimera}, DualPipe only requires that the pipeline stages and micro-batches be divisible by 2, without requiring micro-batches to be divisible by pipeline stages.
In addition, for DualPipe, neither the bubbles nor activation memory will increase as the number of micro-batches grows.

\begin{table}[t]
    \centering
    \setlength{\tabcolsep}{15pt}
    \begin{tabular}{c c c c}
        \toprule
        \textbf{Method} & \textbf{Bubble} & \textbf{Parameter} & \textbf{Activation} \\
        \midrule
        1F1B     & $(PP - 1)(F + B)$       & $1\times$ & $PP$ \\
        ZB1P     & $(PP - 1)(F + B - 2W)$   & $1\times$ & $PP$ \\
        DualPipe (Ours) & $(\frac{PP}{2} - 1)(F\&B + B - 3W)$ & $2\times$ & $PP + 1$ \\
        \bottomrule
    \end{tabular}
    \caption{
        Comparison of pipeline bubbles and memory usage across different pipeline parallel methods. $F$ denotes the execution time of a forward chunk, $B$ denotes the execution time of a full backward chunk, $W$ denotes the execution time of a "backward for weights" chunk, and $F\&B$ denotes the execution time of two mutually overlapped forward and backward chunks.
    }
    \label{tab:dualpipe-bubble}
\end{table}

\subsubsection{Efficient Implementation of Cross-Node All-to-All Communication}

In order to ensure sufficient computational performance for DualPipe, we customize efficient cross-node all-to-all communication kernels (including dispatching and combining) to conserve the number of SMs dedicated to communication.
The implementation of the kernels is co-designed with the MoE gating algorithm and the network topology of our cluster.
To be specific, in our cluster, cross-node GPUs are fully interconnected with IB, and intra-node communications are handled via NVLink.
NVLink offers a bandwidth of 160 GB/s, roughly 3.2 times that of IB (50 GB/s). 
To effectively leverage the different bandwidths of IB and NVLink, we limit each token to be dispatched to at most 4 nodes, thereby reducing IB traffic.
For each token, when its routing decision is made, it will first be transmitted via IB to the GPUs with the same in-node index on its target nodes. 
Once it reaches the target nodes, we will endeavor to ensure that it is instantaneously forwarded via NVLink to specific GPUs that host their target experts, without being blocked by subsequently arriving tokens.
In this way, communications via IB and NVLink are fully overlapped, and each token can efficiently select an average of 3.2 experts per node without incurring additional overhead from NVLink.
This implies that, although \dsviii{} selects only 8 routed experts in practice, it can scale up this number to a maximum of 13 experts (4 nodes $\times$ 3.2 experts/node) while preserving the same communication cost.
Overall, under such a communication strategy, only 20 SMs are sufficient to fully utilize the bandwidths of IB and NVLink.

In detail, we employ the warp specialization technique~\citep{warp-spec} and partition 20 SMs into 10 communication channels.
During the dispatching process, (1) IB sending, (2) IB-to-NVLink forwarding, and (3) NVLink receiving are handled by respective warps. 
The number of warps allocated to each communication task is dynamically adjusted according to the actual workload across all SMs. 
Similarly, during the combining process, (1) NVLink sending, (2) NVLink-to-IB forwarding and accumulation, and (3) IB receiving and accumulation are also handled by dynamically adjusted warps. 
In addition, both dispatching and combining kernels overlap with the computation stream, so we also consider their impact on other SM computation kernels. 
Specifically, we employ customized PTX (Parallel Thread Execution) instructions and auto-tune the communication chunk size, which significantly reduces the use of the L2 cache and the interference to other SMs. 

\subsubsection{Extremely Memory Saving with Minimal Overhead}

In order to reduce the memory footprint during training, we employ the following techniques.

\paragraph{Recomputation of RMSNorm and MLA Up-Projection.}

We recompute all RMSNorm operations and MLA up-projections during back-propagation, thereby eliminating the need to persistently store their output activations.
With a minor overhead, this strategy significantly reduces memory requirements for storing activations. 

\paragraph{Exponential Moving Average in CPU.}
During training, we preserve the Exponential Moving Average (EMA) of the model parameters for early estimation of the model performance after learning rate decay.
The EMA parameters are stored in CPU memory and are updated asynchronously after each training step. 
This method allows us to maintain EMA parameters without incurring additional memory or time overhead.

\paragraph{Shared Embedding and Output Head for Multi-Token Prediction.}
With the DualPipe strategy, we deploy the shallowest layers (including the embedding layer) and deepest layers (including the output head) of the model on the same PP rank. 
This arrangement enables the physical sharing of parameters and gradients, of the shared embedding and output head, between the MTP module and the main model.
This physical sharing mechanism further enhances our memory efficiency. 

\subsection{FP8 Training}

\begin{figure}[!t]
\centering
\includegraphics[width=0.99\linewidth]{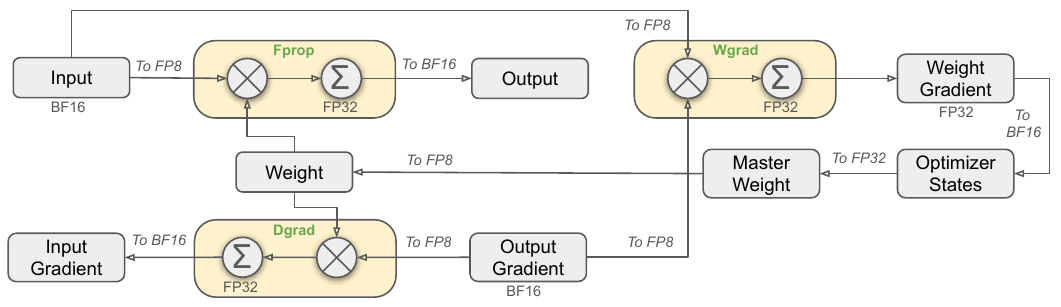}
\caption{
    The overall mixed precision framework with FP8 data format. For clarification, only the \texttt{Linear} operator is illustrated. 
}
\label{fig:fp8_framework}
\end{figure}

Inspired by recent advances in low-precision training~\citep{fp8lm, llm.int8, 8-bit-numerical}, we propose a fine-grained mixed precision framework utilizing the FP8 data format for training \dsviii{}. 
While low-precision training holds great promise, it is often limited by the presence of outliers in activations, weights, and gradients~\citep{massiveoutlier, understandoutlier, scalefp8train}. 
Although significant progress has been made in inference quantization~\citep{xiao2023smoothquant, frantar2022gptq}, there are relatively few studies demonstrating successful application of low-precision techniques in large-scale language model pre-training~\citep{scalefp8train}. 
To address this challenge and effectively extend the dynamic range of the FP8 format, we introduce a fine-grained quantization strategy: tile-wise grouping with $1\times N_c$ elements or block-wise grouping with $N_c\times N_c$ elements. 
The associated dequantization overhead is largely mitigated under our increased-precision accumulation process, a critical aspect for achieving accurate FP8 General Matrix Multiplication~(GEMM). 
Moreover, to further reduce memory and communication overhead in MoE training, we cache and dispatch activations in FP8, while storing low-precision optimizer states in BF16. 
We validate the proposed FP8 mixed precision framework on two model scales similar to \dsvii{}-Lite and \dsvii{}, training for approximately 1 trillion tokens (see more details in Appendix~\ref{app:fp8_cp_bf16}). 
Notably, compared with the BF16 baseline, the relative loss error of our FP8-training model remains consistently below 0.25\%, a level well within the acceptable range of training randomness.

\subsubsection{Mixed Precision Framework}
Building upon widely adopted techniques in low-precision training~\citep{bf16train, fp16train}, we propose a mixed precision framework for FP8 training. 
In this framework, most compute-density operations are conducted in FP8, while a few key operations are strategically maintained in their original data formats to balance training efficiency and numerical stability. 
The overall framework is illustrated in Figure~\ref{fig:fp8_framework}.

Firstly, in order to accelerate model training, the majority of core computation kernels, i.e., GEMM operations, are implemented in FP8 precision. 
These GEMM operations accept FP8 tensors as inputs and produce outputs in BF16 or FP32. 
As depicted in Figure~\ref{fig:fp8_framework}, all three GEMMs associated with the \texttt{Linear} operator, namely \texttt{Fprop} (forward pass), \texttt{Dgrad} (activation backward pass), and \texttt{Wgrad} (weight backward pass), are executed in FP8. 
This design theoretically doubles the computational speed compared with the original BF16 method. 
Additionally, the FP8 \texttt{Wgrad} GEMM allows activations to be stored in FP8 for use in the backward pass. 
This significantly reduces memory consumption.

Despite the efficiency advantage of the FP8 format, certain operators still require a higher precision due to their sensitivity to low-precision computations. 
Besides, some low-cost operators can also utilize a higher precision with a negligible overhead to the overall training cost. 
For this reason, after careful investigations, we maintain the original precision (e.g., BF16 or FP32) for the following components: the embedding module, the output head, MoE gating modules, normalization operators, and attention operators. 
These targeted retentions of high precision ensure stable training dynamics for \dsviii{}.
To further guarantee numerical stability, we store the master weights, weight gradients, and optimizer states in higher precision. 
While these high-precision components incur some memory overheads, their impact can be minimized through efficient sharding across multiple DP ranks in our distributed training system.

\begin{figure}[!t]
\centering
\includegraphics[width=0.95\linewidth]{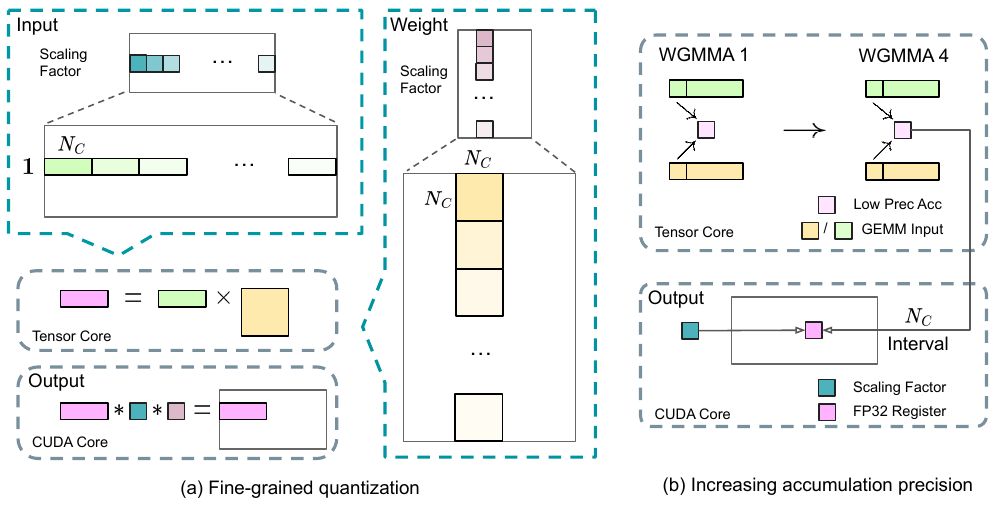}
\caption{
    (a) We propose a fine-grained quantization method to mitigate quantization errors caused by feature outliers; for illustration simplicity, only \texttt{Fprop} is illustrated. 
    (b) In conjunction with our quantization strategy, we improve the FP8 GEMM precision by promoting to CUDA Cores at an interval of $N_C=128$ elements MMA for the high-precision accumulation. 
}
\label{fig:fp8_quantization}
\end{figure}

\subsubsection{Improved Precision from Quantization and Multiplication}
Based on our mixed precision FP8 framework, we introduce several strategies to enhance low-precision training accuracy, focusing on both the quantization method and the multiplication process.

\paragraph{Fine-Grained Quantization.}
In low-precision training frameworks, overflows and underflows are common challenges due to the limited dynamic range of the FP8 format, which is constrained by its reduced exponent bits. 
As a standard practice, the input distribution is aligned to the representable range of the FP8 format by scaling the maximum absolute value of the input tensor to the maximum representable value of FP8~\citep{fp16train}. 
This method makes low-precision training highly sensitive to activation outliers, which can heavily degrade quantization accuracy.
To solve this, we propose a fine-grained quantization method that applies scaling at a more granular level. 
As illustrated in Figure~\ref{fig:fp8_quantization} (a), (1) for activations, we group and scale elements on a \texttt{1x128} tile basis (i.e., per token per 128 channels); and (2) for weights, we group and scale elements on a \texttt{128x128} block basis (i.e., per 128 input channels per 128 output channels).
This approach ensures that the quantization process can better accommodate outliers by adapting the scale according to smaller groups of elements. 
In Appendix~\ref{app:fp8_blockwise}, we further discuss the training instability when we group and scale activations on a block basis in the same way as weights quantization.

One key modification in our method is the introduction of per-group scaling factors along the inner dimension of GEMM operations. 
This functionality is not directly supported in the standard FP8 GEMM. 
However, combined with our precise FP32 accumulation strategy, it can be efficiently implemented. 

Notably, our fine-grained quantization strategy is highly consistent with the idea of microscaling formats~\citep{rouhani2023microscaling}, while the Tensor Cores of NVIDIA next-generation GPUs (Blackwell series) have announced the support for microscaling formats with smaller quantization granularity~\citep{nvidia_tensor_cores}. 
We hope our design can serve as a reference for future work to keep pace with the latest GPU architectures.

\paragraph{Increasing Accumulation Precision.}
Low-precision GEMM operations often suffer from underflow issues, and their accuracy largely depends on high-precision accumulation, which is commonly performed in an FP32 precision~\citep{bf16train, fp16train}.
However, we observe that the accumulation precision of FP8 GEMM on NVIDIA H800 GPUs is limited to retaining around 14 bits, which is significantly lower than FP32 accumulation precision. 
This problem will become more pronounced when the inner dimension \texttt{K} is large~\citep{switchback}, a typical scenario in large-scale model training where the batch size and model width are increased. 
Taking GEMM operations of two random matrices with \texttt{K} = 4096 for example, in our preliminary test, the limited accumulation precision in Tensor Cores results in a maximum relative error of nearly 2\%. 
Despite these problems, the limited accumulation precision is still the default option in a few FP8 frameworks~\citep{transformerengine}, severely constraining the training accuracy. 

In order to address this issue, we adopt the strategy of promotion to CUDA Cores for higher precision~\citep{Thakkar_CUTLASS_2023}. 
The process is illustrated in Figure~\ref{fig:fp8_quantization} (b). 
To be specific, during MMA (Matrix Multiply-Accumulate) execution on Tensor Cores, intermediate results are accumulated using the limited bit width. 
Once an interval of $N_C$ is reached, these partial results will be copied to FP32 registers on CUDA Cores, where full-precision FP32 accumulation is performed.
As mentioned before, our fine-grained quantization applies per-group scaling factors along the inner dimension \texttt{K}. 
These scaling factors can be efficiently multiplied on the CUDA Cores as the dequantization process with minimal additional computational cost.

It is worth noting that this modification reduces the WGMMA (Warpgroup-level Matrix Multiply-Accumulate) instruction issue rate for a single warpgroup. However, on the H800 architecture, it is typical for two WGMMA to persist concurrently: while one warpgroup performs the promotion operation, the other is able to execute the MMA operation. This design enables overlapping of the two operations, maintaining high utilization of Tensor Cores.
Based on our experiments, setting $N_C=128$ elements, equivalent to 4 WGMMAs, represents the minimal accumulation interval that can significantly improve precision without introducing substantial overhead.

\paragraph{Mantissa over Exponents.} 
In contrast to the hybrid FP8 format adopted by prior work \citep{transformerengine, fp8lm, hfp8}, which uses \texttt{E4M3} (4-bit exponent and 3-bit mantissa) in \texttt{Fprop} and \texttt{E5M2} (5-bit exponent and 2-bit mantissa) in \texttt{Dgrad} and \texttt{Wgrad}, we adopt the \texttt{E4M3} format on all tensors for higher precision. 
We attribute the feasibility of this approach to our fine-grained quantization strategy, i.e., tile and block-wise scaling. 
By operating on smaller element groups, our methodology effectively shares exponent bits among these grouped elements, mitigating the impact of the limited dynamic range. 

\paragraph{Online Quantization.}
Delayed quantization is employed in tensor-wise quantization frameworks~\citep{transformerengine, fp8lm}, which maintains a history of the maximum absolute values across prior iterations to infer the current value. 
In order to ensure accurate scales and simplify the framework, we calculate the maximum absolute value online for each \texttt{1x128} activation tile or \texttt{128x128} weight block. 
Based on it, we derive the scaling factor and then quantize the activation or weight online into the FP8 format.

\subsubsection{Low-Precision Storage and Communication}
In conjunction with our FP8 training framework, we further reduce the memory consumption and communication overhead by compressing cached activations and optimizer states into lower-precision formats.

\paragraph{Low-Precision Optimizer States.}
We adopt the BF16 data format instead of FP32 to track the first and second moments in the AdamW~\citep{adamW} optimizer, without incurring observable performance degradation. 
However, the master weights (stored by the optimizer) and gradients (used for batch size accumulation) are still retained in FP32 to ensure numerical stability throughout training.

\paragraph{Low-Precision Activation.}
As illustrated in Figure~\ref{fig:fp8_framework}, the \texttt{Wgrad} operation is performed in FP8. 
To reduce the memory consumption, it is a natural choice to cache activations in FP8 format for the backward pass of the \texttt{Linear} operator.
However, special considerations are taken on several operators for low-cost high-precision training:
\begin{quote}
\textbf{(1) Inputs of the \texttt{Linear} after the attention operator.} These activations are also used in the backward pass of the attention operator, which makes it sensitive to precision. We adopt a customized \texttt{E5M6} data format exclusively for these activations. Additionally, these activations will be converted from an \texttt{1x128} quantization tile to an \texttt{128x1} tile in the backward pass. To avoid introducing extra quantization error, all the scaling factors are round scaled, i.e., integral power of 2.

\textbf{(2) Inputs of the SwiGLU operator in MoE.} To further reduce the memory cost, we cache the inputs of the SwiGLU operator and recompute its output in the backward pass. These activations are also stored in FP8 with our fine-grained quantization method, striking a balance between memory efficiency and computational accuracy.
\end{quote}

\paragraph{Low-Precision Communication.}
Communication bandwidth is a critical bottleneck in the training of MoE models. 
To alleviate this challenge, we quantize the activation before MoE up-projections into FP8 and then apply \texttt{dispatch} components, which is compatible with FP8 \texttt{Fprop} in MoE up-projections. Like the inputs of the \texttt{Linear} after the attention operator, scaling factors for this activation are integral power of 2.
A similar strategy is applied to the activation gradient before MoE down-projections. 
For both the forward and backward \texttt{combine} components, we retain them in BF16 to preserve training precision in critical parts of the training pipeline.

\subsection{Inference and Deployment}
\label{sec:inference_deployment}

We deploy \dsviii{} on the H800 cluster, where GPUs within each node are interconnected using NVLink, and all GPUs across the cluster are fully interconnected via IB. 
To simultaneously ensure both the Service-Level Objective (SLO) for online services and high throughput, we employ the following deployment strategy that separates the \textit{prefilling} and \textit{decoding} stages.

\subsubsection{Prefilling}

The minimum deployment unit of the prefilling stage consists of 4 nodes with 32 GPUs. 
The \texttt{attention} part employs 4-way Tensor Parallelism (TP4) with Sequence Parallelism (SP), combined with 8-way Data Parallelism (DP8).
Its small TP size of 4 limits the overhead of TP communication. 
For the \texttt{MoE} part, we use 32-way Expert Parallelism (EP32), which ensures that each expert processes a sufficiently large batch size, thereby enhancing computational efficiency. 
For the \texttt{MoE} all-to-all communication, we use the same method as in training: first transferring tokens across nodes via IB, and then forwarding among the intra-node GPUs via NVLink. 
In particular, we use 1-way Tensor Parallelism for the dense MLPs in shallow layers to save TP communication.

To achieve load balancing among different experts in the \texttt{MoE} part, we need to ensure that each GPU processes approximately the same number of tokens.  
To this end, we introduce a deployment strategy of \textit{redundant experts}, which duplicates high-load experts and deploys them redundantly. 
The high-load experts are detected based on statistics collected during the online deployment and are adjusted periodically (e.g., every 10 minutes). 
After determining the set of redundant experts, we carefully rearrange experts among GPUs within a node based on the observed loads, striving to balance the load across GPUs as much as possible without increasing the cross-node all-to-all communication overhead. 
For the deployment of \dsviii{}, we set 32 redundant experts for the prefilling stage. 
For each GPU, besides the original 8 experts it hosts, it will also host one additional redundant expert.

Furthermore, in the prefilling stage, to improve the throughput and hide the overhead of all-to-all and TP communication, we simultaneously process two micro-batches with similar computational workloads, overlapping the \texttt{attention} and \texttt{MoE} of one micro-batch with the \texttt{dispatch} and \texttt{combine} of another. 

Finally, we are exploring a \textit{dynamic redundancy} strategy for experts, where each GPU hosts more experts (e.g., 16 experts), but only 9 will be activated during each inference step. 
Before the all-to-all operation at each layer begins, we compute the globally optimal routing scheme on the fly. 
Given the substantial computation involved in the prefilling stage, the overhead of computing this routing scheme is almost negligible.

\subsubsection{Decoding}

During decoding, we treat the shared expert as a routed one. 
From this perspective, each token will select 9 experts during routing, where the shared expert is regarded as a heavy-load one that will always be selected. 
The minimum deployment unit of the decoding stage consists of 40 nodes with 320 GPUs. 
The \texttt{attention} part employs TP4 with SP, combined with DP80, while the \texttt{MoE} part uses EP320. 
For the \texttt{MoE} part, each GPU hosts only one expert, and 64 GPUs are responsible for hosting redundant experts and shared experts.
All-to-all communication of the \texttt{dispatch} and \texttt{combine} parts is performed via direct point-to-point transfers over IB to achieve low latency. 
Additionally, we leverage the IBGDA~\citep{nvidia_ibgda} technology to further minimize latency and enhance communication efficiency.

Similar to prefilling, we periodically determine the set of redundant experts in a certain interval, based on the statistical expert load from our online service. 
However, we do not need to rearrange experts since each GPU only hosts one expert. 
We are also exploring the \textit{dynamic redundancy} strategy for decoding. 
However, this requires more careful optimization of the algorithm that computes the globally optimal routing scheme and the fusion with the \texttt{dispatch} kernel to reduce overhead.

Additionally, to enhance throughput and hide the overhead of all-to-all communication, we are also exploring processing two micro-batches with similar computational workloads simultaneously in the decoding stage. 
Unlike prefilling, \texttt{attention} consumes a larger portion of time in the decoding stage. Therefore, we overlap the \texttt{attention} of one micro-batch with the \texttt{dispatch+MoE+combine} of another.
In the decoding stage, the batch size per expert is relatively small (usually within 256 tokens), and the bottleneck is memory access rather than computation. 
Since the \texttt{MoE} part only needs to load the parameters of one expert, the memory access overhead is minimal, so using fewer SMs will not significantly affect the overall performance.
Therefore, to avoid impacting the computation speed of the \texttt{attention} part, we can allocate only a small portion of SMs to \texttt{dispatch+MoE+combine}. 

\subsection{Suggestions on Hardware Design}
\label{fp8_hardware_design}

Based on our implementation of the all-to-all communication and FP8 training scheme, we propose the following suggestions on chip design to AI hardware vendors.

\subsubsection{Communication Hardware}

In \dsviii{}, we implement the overlap between computation and communication to hide the communication latency during computation. 
This significantly reduces the dependency on communication bandwidth compared to serial computation and communication. 
However, the current communication implementation relies on expensive SMs (e.g., we allocate 20 out of the 132 SMs available in the H800 GPU for this purpose), which will limit the computational throughput.
Moreover, using SMs for communication results in significant inefficiencies, as tensor cores remain entirely under-utilized.

Currently, the SMs primarily perform the following tasks for all-to-all communication:
\begin{itemize}[topsep=0pt]
    \item 
    \textbf{Forwarding data} between the IB (InfiniBand) and NVLink domain while aggregating IB traffic destined for multiple GPUs within the same node from a single GPU.
    \item 
    \textbf{Transporting data} between RDMA buffers (registered GPU memory regions) and input/output buffers.
    \item 
    \textbf{Executing \texttt{reduce} operations} for \texttt{all-to-all} \texttt{combine}.
    \item
    \textbf{Managing fine-grained memory layout} during chunked data transferring to multiple experts across the IB and NVLink domain.
\end{itemize}

We aspire to see future vendors developing hardware that offloads these communication tasks from the valuable computation unit SM, serving as a GPU co-processor or a network co-processor like NVIDIA SHARP~\cite{nvsharp}. 
Furthermore, to reduce application programming complexity, we aim for this hardware to unify the IB (scale-out) and NVLink (scale-up) networks from the perspective of the computation units. 
With this unified interface, computation units can easily accomplish operations such as \texttt{read}, \texttt{write}, \texttt{multicast}, and \texttt{reduce} across the entire IB-NVLink-unified domain via submitting communication requests based on simple primitives.

\subsubsection{Compute Hardware}

\paragraph{Higher FP8 GEMM Accumulation Precision in Tensor Cores.}
In the current Tensor Core implementation of the NVIDIA Hopper architecture, FP8 GEMM suffers from limited accumulation precision. After aligning 32 mantissa products by right-shifting based on the maximum exponent, the Tensor Core only uses the highest 14 bits of each mantissa product for addition, and truncates bits exceeding this range. The accumulation of addition results into registers also employs 14-bit precision. Our implementation partially mitigates the limitation by accumulating the addition results of 128 FP8$\times$FP8 multiplications into registers with FP32 precision in the CUDA core. 
Although helpful in achieving successful FP8 training, it is merely a compromise due to the Hopper architecture's hardware deficiency in FP8 GEMM accumulation precision.
Future chips need to adopt higher precision.

\paragraph{Support for Tile- and Block-Wise Quantization.}
Current GPUs only support per-tensor quantization, lacking the native support for fine-grained quantization like our tile- and block-wise quantization.
In the current implementation, when the $N_C$ interval is reached, the partial results will be copied from Tensor Cores to CUDA cores, multiplied by the scaling factors, and added to FP32 registers on CUDA cores.
Although the dequantization overhead is significantly mitigated combined with our precise FP32 accumulation strategy, the frequent data movements between Tensor Cores and CUDA cores still limit the computational efficiency. 
Therefore, we recommend future chips to support fine-grained quantization by enabling Tensor Cores to receive scaling factors and implement MMA with group scaling. 
In this way, the whole partial sum accumulation and dequantization can be completed directly inside Tensor Cores until the final result is produced, avoiding frequent data movements.

\paragraph{Support for Online Quantization.}
The current implementations struggle to effectively support online quantization, despite its effectiveness demonstrated in our research. 
In the existing process, we need to read 128 BF16 activation values (the output of the previous computation) from HBM (High Bandwidth Memory) for quantization, and the quantized FP8 values are then written back to HBM, only to be read again for MMA. 
To address this inefficiency, we recommend that future chips integrate FP8 cast and TMA (Tensor Memory Accelerator) access into a single fused operation, so quantization can be completed during the transfer of activations from global memory to shared memory, avoiding frequent memory reads and writes. 
We also recommend supporting a warp-level cast instruction for speedup, which further facilitates the better fusion of layer normalization and FP8 cast.
Alternatively, a near-memory computing approach can be adopted, where compute logic is placed near the HBM. 
In this case, BF16 elements can be cast to FP8 directly as they are read from HBM into the GPU, reducing off-chip memory access by roughly 50\%. 

\paragraph{Support for Transposed GEMM Operations.}
The current architecture makes it cumbersome to fuse matrix transposition with GEMM operations. 
In our workflow, activations during the forward pass are quantized into \texttt{1x128} FP8 tiles and stored. 
During the backward pass, the matrix needs to be read out, dequantized, transposed, re-quantized into \texttt{128x1} tiles, and stored in HBM. 
To reduce memory operations, we recommend future chips to enable direct transposed reads of matrices from shared memory before MMA operation, for those precisions required in both training and inference. Combined with the fusion of FP8 format conversion and TMA access, this enhancement will significantly streamline the quantization workflow.

\section{Pre-Training}
\label{sec:pre-training}

\subsection{Data Construction}

Compared with \dsvii{}, we optimize the pre-training corpus by enhancing the ratio of mathematical and programming samples, while expanding multilingual coverage beyond English and Chinese. 
Also, our data processing pipeline is refined to minimize redundancy while maintaining corpus diversity. 
Inspired by \citet{Ding2024FewerTI}, we implement the document packing method for data integrity but do not incorporate cross-sample attention masking during training. 
Finally, the training corpus for \dsviii{} consists of 14.8T high-quality and diverse tokens in our tokenizer. 

In the training process of DeepSeekCoder-V2~\citep{dscodervii}, we observe that the Fill-in-Middle~(FIM) strategy does not compromise the next-token prediction capability while enabling the model to accurately predict middle text based on contextual cues. 
In alignment with DeepSeekCoder-V2, we also incorporate the FIM strategy in the pre-training of \dsviii{}. 
To be specific, we employ the Prefix-Suffix-Middle (PSM) framework to structure data as follows:
\begin{align}
\texttt{<|fim\_begin|>}f_{\text{pre}}\texttt{<|fim\_hole|>}f_{\text{suf}}\texttt{<|fim\_end|>}f_{\text{middle}}\texttt{<|eos\_token|>} . \nonumber
\end{align}
This structure is applied at the document level as a part of the pre-packing process. 
The FIM strategy is applied at a rate of 0.1, consistent with the PSM framework.

The tokenizer for \dsviii{} employs Byte-level BPE~\citep{shibata1999byte} with an extended vocabulary of 128K tokens. 
The pretokenizer and training data for our tokenizer are modified to optimize multilingual compression efficiency. 
In addition, compared with \dsvii{}, the new pretokenizer introduces tokens that combine punctuations and line breaks. 
However, this trick may introduce the token boundary bias~\citep{tokenboundary} when the model processes multi-line prompts without terminal line breaks, particularly for few-shot evaluation prompts. 
To address this issue, we randomly split a certain proportion of such combined tokens during training, which exposes the model to a wider array of special cases and mitigates this bias. 

\subsection{Hyper-Parameters}

\paragraph{Model Hyper-Parameters.}
We set the number of Transformer layers to 61 and the hidden dimension to 7168. 
All learnable parameters are randomly initialized with a standard deviation of 0.006.
In \dsattn{}, we set the number of attention heads $n_h$ to 128 and the per-head dimension $d_h$ to 128. 
The KV compression dimension $d_c$ is set to 512, and the query compression dimension $d_c^{\prime}$ is set to 1536. 
For the decoupled queries and key, we set the per-head dimension $d_h^R$ to 64. 
We substitute all FFNs except for the first three layers with MoE layers. 
Each MoE layer consists of 1 shared expert and 256 routed experts, where the intermediate hidden dimension of each expert is 2048. 
Among the routed experts, 8 experts will be activated for each token, and each token will be ensured to be sent to at most 4 nodes. 
The multi-token prediction depth $D$ is set to 1, i.e., besides the exact next token, each token will predict one additional token. 
As \dsvii{}, \dsviii{} also employs additional RMSNorm layers after the compressed latent vectors, and multiplies additional scaling factors at the width bottlenecks. 
Under this configuration, \dsviii{} comprises 671B total parameters, of which 37B are activated for each token. 

\paragraph{Training Hyper-Parameters.}
We employ the AdamW optimizer~\citep{adamW} with hyper-parameters set to $\beta_1=0.9$, $\beta_2=0.95$, and $\mathrm{weight\_decay}=0.1$. 
We set the maximum sequence length to 4K during pre-training, and pre-train \dsviii{} on 14.8T tokens. 
As for the learning rate scheduling, we first linearly increase it from 0 to $2.2 \times 10^{-4}$ during the first 2K steps. 
Then, we keep a constant learning rate of $2.2 \times 10^{-4}$ until the model consumes 10T training tokens. 
Subsequently, we gradually decay the learning rate to $2.2 \times 10^{-5}$ in 4.3T tokens, following a cosine decay curve. 
During the training of the final 500B tokens, we keep a constant learning rate of $2.2 \times 10^{-5}$ in the first 333B tokens, and switch to another constant learning rate of $7.3 \times 10^{-6}$ in the remaining 167B tokens. 
The gradient clipping norm is set to 1.0.
We employ a batch size scheduling strategy, where the batch size is gradually increased from 3072 to 15360 in the training of the first 469B tokens, and then keeps 15360 in the remaining training. 
We leverage pipeline parallelism to deploy different layers of a model on different GPUs, and for each layer, the routed experts will be uniformly deployed on 64 GPUs belonging to 8 nodes. 
As for the node-limited routing, each token will be sent to at most 4 nodes (i.e., $M=4$). 
For auxiliary-loss-free load balancing, we set the bias update speed $\gamma$ to 0.001 for the first 14.3T tokens, and to 0.0 for the remaining 500B tokens. 
For the balance loss, we set $\alpha$ to 0.0001, just to avoid extreme imbalance within any single sequence. 
The MTP loss weight $\lambda$ is set to 0.3 for the first 10T tokens, and to 0.1 for the remaining 4.8T tokens. 

\begin{figure}[h]
\centering
\includegraphics[width=0.98\linewidth]{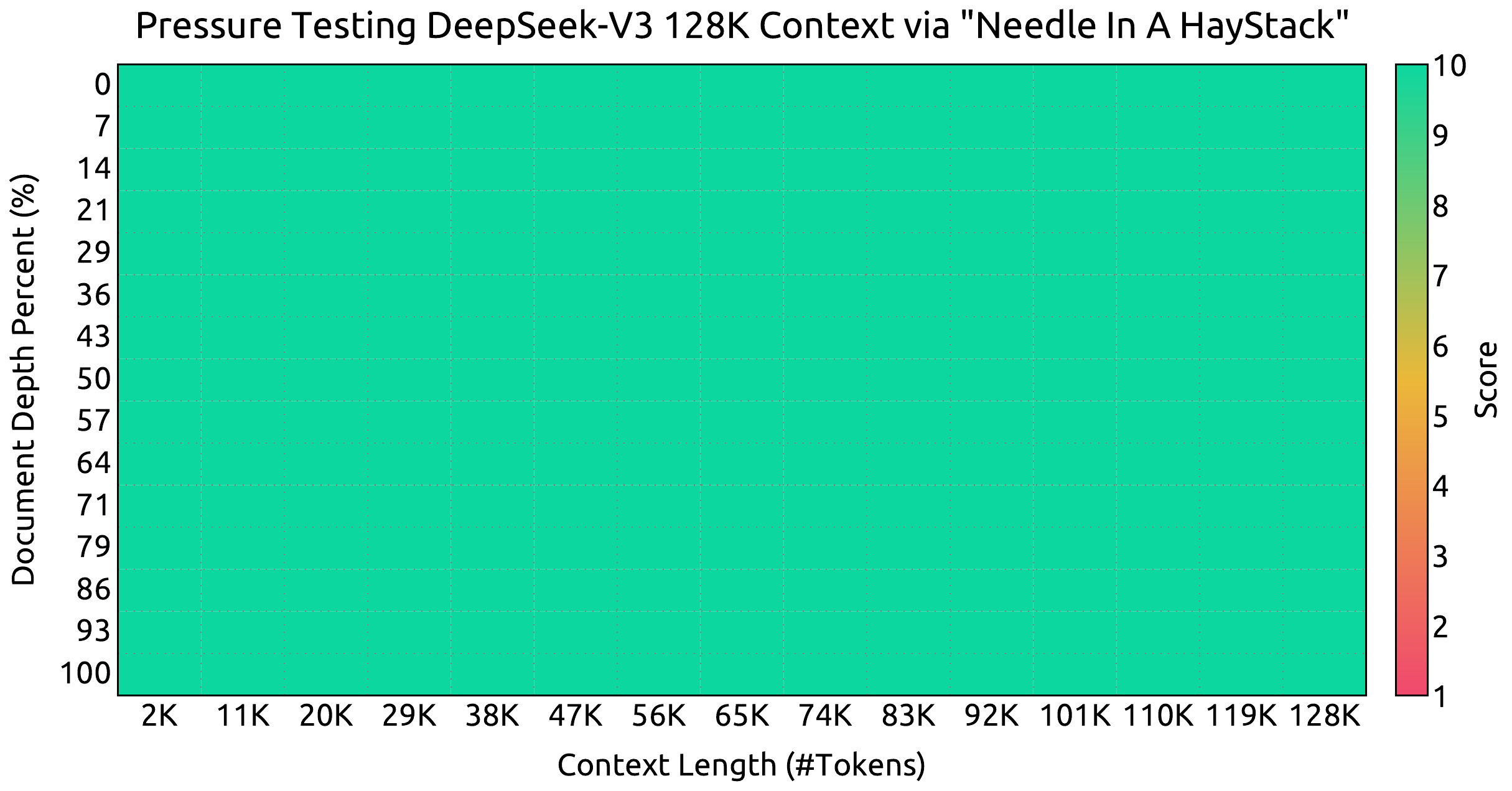}
\caption{
Evaluation results on the ''Needle In A Haystack'' (NIAH) tests. 
\dsviii{} performs well across all context window lengths up to 128K. 
}
\label{fig:long_context}
\end{figure}

\subsection{Long Context Extension}

We adopt a similar approach to \dsvii{}~\citep{dsvii} to enable long context capabilities in \dsviii{}. 
After the pre-training stage, we apply YaRN~\citep{peng2023yarn} for context extension and perform two additional training phases, each comprising 1000 steps, to progressively expand the context window from 4K to 32K and then to 128K.
The YaRN configuration is consistent with that used in \dsvii{}, being applied exclusively to the decoupled shared key $\mathbf{k}^R_t$.
The hyper-parameters remain identical across both phases, with the scale $s = 40$, $\alpha = 1$, $\beta = 32$, and the scaling factor $\sqrt{t} = 0.1 \ln{s} + 1$.
In the first phase, the sequence length is set to 32K, and the batch size is 1920. 
During the second phase, the sequence length is increased to 128K, and the batch size is reduced to 480. 
The learning rate for both phases is set to $7.3 \times 10^{-6}$, matching the final learning rate from the pre-training stage.

Through this two-phase extension training, \dsviii{} is capable of handling inputs up to 128K in length while maintaining strong performance. 
Figure~\ref{fig:long_context} illustrates that \dsviii{}, following supervised fine-tuning, achieves notable performance on the "Needle In A Haystack" (NIAH) test, demonstrating consistent robustness across context window lengths up to 128K.

\subsection{Evaluations}

\subsubsection{Evaluation Benchmarks}

The base model of \dsviii{} is pretrained on a multilingual corpus with English and Chinese constituting the majority, so we evaluate its performance on a series of benchmarks primarily in English and Chinese, as well as on a multilingual benchmark.
Our evaluation is based on our internal evaluation framework integrated in our HAI-LLM framework. 
Considered benchmarks are categorized and listed as follows, where \underline{underlined} benchmarks are in Chinese and \uuline{double-underlined} benchmarks are multilingual ones:

\textbf{Multi-subject multiple-choice} datasets include MMLU \citep{mmlu}, MMLU-Redux \citep{mmlu_redux}, MMLU-Pro \citep{mmlu_pro}, \uuline{MMMLU} \citep{mmmlu}, \underline{C-Eval} \citep{ceval}, and \underline{CMMLU} \citep{cmmlu}.

\textbf{Language understanding and reasoning} datasets include HellaSwag \citep{hellaswag}, PIQA \citep{piqa}, ARC \citep{arc}, and BigBench Hard (BBH) \citep{bbh}.

\textbf{Closed-book question answering} datasets include TriviaQA \citep{joshi-etal-2017-triviaqa} and NaturalQuestions \citep{naturalquestions}.

\textbf{Reading comprehension} datasets include RACE \cite{race}, DROP \citep{drop}, \underline{C3} \citep{sun2019investigating}, and \underline{CMRC} \citep{cui-etal-2019-span}.

\textbf{Reference disambiguation} datasets include \underline{CLUEWSC} \citep{clue} and WinoGrande \cite{sakaguchi2019winogrande}.

\textbf{Language modeling} datasets include Pile \citep{pile}.

\textbf{Chinese understanding and culture} datasets include \underline{CCPM} \citep{li2021ccpm}.

\textbf{Math} datasets include GSM8K~\citep{gsm8k}, MATH~\citep{hendrycks2021measuring}, MGSM \citep{mgsm}, and \underline{CMath} \citep{wei2023cmath}.

\textbf{Code} datasets include HumanEval~\citep{codex}, LiveCodeBench-Base (0801-1101) \citep{livecodebench}, MBPP~\citep{mbpp}, and CRUXEval~\citep{gu2024cruxeval}.

\textbf{Standardized exams} include \underline{AGIEval} \citep{agieval}. 
Note that AGIEval includes both English and Chinese subsets.

Following our previous work~\citep{dsvi,dsvii}, we adopt perplexity-based evaluation for datasets including HellaSwag, PIQA, WinoGrande, RACE-Middle, RACE-High, MMLU, MMLU-Redux, MMLU-Pro, MMMLU, ARC-Easy, ARC-Challenge, C-Eval, CMMLU, C3, and CCPM, and adopt generation-based evaluation for TriviaQA, NaturalQuestions, DROP, MATH, GSM8K, MGSM, HumanEval, MBPP, LiveCodeBench-Base, CRUXEval, BBH, AGIEval, CLUEWSC, CMRC, and CMath. 
In addition, we perform language-modeling-based evaluation for Pile-test and use Bits-Per-Byte~(BPB) as the metric to guarantee fair comparison among models using different tokenizers. 

\begin{table}[!h]
    \centering
    \footnotesize
    \setlength{\tabcolsep}{4.5pt}
    \begin{tabular}{@{}c l c | c | c c | c@{}}
    \toprule
    & \multirow{2}{*}{\centering \textbf{Benchmark {\tiny (Metric)}}} & \multirow{2}{*}{\textbf{\# Shots}} & \textbf{\dsvii{}} & \textbf{Qwen2.5} & \textbf{LLaMA-3.1} & \textbf{\dsviii{}} \\
    & & & \textbf{Base} & \textbf{72B Base} & \textbf{405B Base} & \textbf{Base} \\
    \midrule
    & Architecture & - & MoE & Dense & Dense & MoE \\
    & \# Activated Params & - & 21B & 72B & 405B & 37B \\
    & \# Total Params & - & 236B & 72B & 405B & 671B \\
    \midrule
    \multirow{16}{*}{English} & Pile-test {\tiny (BPB)} & - & 0.606 & 0.638 & \textbf{0.542} & 0.548 \\
    & BBH {\tiny (EM)} & 3-shot & 78.8 & 79.8 & 82.9 & \textbf{87.5} \\
    & MMLU {\tiny (EM)} & 5-shot & 78.4 & 85.0 & 84.4 & \textbf{87.1} \\
    & MMLU-Redux {\tiny (EM)} & 5-shot & 75.6 & 83.2 & 81.3 & \textbf{86.2} \\
    & MMLU-Pro {\tiny (EM)} & 5-shot & 51.4 & 58.3 & 52.8 & \textbf{64.4} \\
    & DROP {\tiny (F1)} & 3-shot & 80.4 & 80.6 & 86.0 & \textbf{89.0} \\
    & ARC-Easy {\tiny (EM)} & 25-shot & 97.6 & 98.4 & 98.4 & \textbf{98.9} \\
    & ARC-Challenge {\tiny (EM)} & 25-shot & 92.2 & 94.5 & \textbf{95.3} & \textbf{95.3} \\
    & HellaSwag {\tiny (EM)} & 10-shot & 87.1 & 84.8 & \textbf{89.2} & \textbf{88.9} \\
    & PIQA {\tiny (EM)} & 0-shot & 83.9 & 82.6 & \textbf{85.9} & 84.7 \\
    & WinoGrande {\tiny (EM)} & 5-shot & \textbf{86.3} & 82.3 & 85.2 & 84.9 \\
    & RACE-Middle {\tiny (EM)} & 5-shot & 73.1 & 68.1 & \textbf{74.2} & 67.1 \\
    & RACE-High {\tiny (EM)} & 5-shot & 52.6 & 50.3 & \textbf{56.8} & 51.3 \\
    & TriviaQA {\tiny (EM)} & 5-shot & 80.0 & 71.9 & \textbf{82.7} & \textbf{82.9} \\
    & NaturalQuestions {\tiny (EM)} & 5-shot & 38.6 & 33.2 & \textbf{41.5} & 40.0 \\
    & AGIEval {\tiny (EM)} & 0-shot & 57.5 & 75.8 & 60.6 & \textbf{79.6} \\
    \midrule
    \multirow{4}{*}{Code} & HumanEval {\tiny (Pass@1)} & 0-shot & 43.3 & 53.0 & 54.9 & \textbf{65.2} \\
    & MBPP {\tiny (Pass@1)} & 3-shot & 65.0 & 72.6 & 68.4 & \textbf{75.4} \\
    & LiveCodeBench-Base {\tiny (Pass@1)} & 3-shot & 11.6 & 12.9 & 15.5 & \textbf{19.4} \\
    & CRUXEval-I {\tiny (EM)} & 2-shot & 52.5 & 59.1 & 58.5 & \textbf{67.3} \\
    & CRUXEval-O {\tiny (EM)} & 2-shot & 49.8 & 59.9 & 59.9 & \textbf{69.8} \\
    \midrule
    \multirow{3}{*}{Math} & GSM8K {\tiny (EM)} & 8-shot & 81.6 & 88.3 & 83.5 & \textbf{89.3} \\
    & MATH {\tiny (EM)} & 4-shot & 43.4 & 54.4 & 49.0 & \textbf{61.6} \\
    & MGSM {\tiny (EM)} & 8-shot & 63.6 & 76.2 & 69.9 & \textbf{79.8} \\
    & CMath {\tiny (EM)} & 3-shot & 78.7 & 84.5 & 77.3 & \textbf{90.7} \\
    \midrule
    \multirow{7}{*}{Chinese} & CLUEWSC {\tiny (EM)} & 5-shot & 82.0 & 82.5 & \textbf{83.0} & \textbf{82.7} \\
    & C-Eval {\tiny (EM)} & 5-shot & 81.4 & 89.2 & 72.5 & \textbf{90.1} \\
    & CMMLU {\tiny (EM)} & 5-shot & 84.0 & \textbf{89.5} & 73.7 & 88.8 \\
    & CMRC {\tiny (EM)} & 1-shot & \textbf{77.4} & 75.8 & 76.0 & 76.3 \\
    & C3 {\tiny (EM)} & 0-shot & 77.4 & 76.7 & \textbf{79.7} & 78.6 \\
    & CCPM {\tiny (EM)} & 0-shot & \textbf{93.0} & 88.5 & 78.6 & 92.0 \\
    \midrule
    \multirow{1}{*}{Multilingual} & MMMLU-non-English {\tiny (EM)} & 5-shot & 64.0 & 74.8 & 73.8 & \textbf{79.4} \\
    \bottomrule
    \end{tabular}
    \caption{
        Comparison among \dsviii{}-Base and other representative open-source base models.
        All models are evaluated in our internal framework and share the same evaluation setting.
        Scores with a gap not exceeding 0.3 are considered to be at the same level. 
        \dsviii{}-Base achieves the best performance on most benchmarks, especially on math and code tasks. 
    }
    \label{tab:main}
\end{table}

\subsubsection{Evaluation Results}

In Table~\ref{tab:main}, we compare the base model of \dsviii{} with the state-of-the-art open-source base models, including \dsvii{}-Base~\citep{dsvii} (our previous release), Qwen2.5 72B Base~\citep{qwen2_5}, and LLaMA-3.1 405B Base~\citep{llama3_1_405b}.
We evaluate all these models with our internal evaluation framework, and ensure that they share the same evaluation setting. 
Note that due to the changes in our evaluation framework over the past months, the performance of \dsvii{}-Base exhibits a slight difference from our previously reported results.
Overall, \dsviii{}-Base comprehensively outperforms \dsvii{}-Base and Qwen2.5 72B Base, and surpasses LLaMA-3.1 405B Base in the majority of benchmarks, essentially becoming the strongest open-source model.  

From a more detailed perspective, we compare \dsviii{}-Base with the other open-source base models individually.
(1) 
Compared with \dsvii{}-Base, due to the improvements in our model architecture, the scale-up of the model size and training tokens, and the enhancement of data quality, \dsviii{}-Base achieves significantly better performance as expected.
(2)
Compared with Qwen2.5 72B Base, the state-of-the-art Chinese open-source model, with only half of the activated parameters, \dsviii{}-Base also demonstrates remarkable advantages, especially on English, multilingual, code, and math benchmarks. 
As for Chinese benchmarks, except for CMMLU, a Chinese multi-subject multiple-choice task, \dsviii{}-Base also shows better performance than Qwen2.5 72B. 
(3)
Compared with LLaMA-3.1 405B Base, the largest open-source model with 11 times the activated parameters, \dsviii{}-Base also exhibits much better performance on multilingual, code, and math benchmarks. 
As for English and Chinese language benchmarks, \dsviii{}-Base shows competitive or better performance, and is especially good on BBH, MMLU-series, DROP, C-Eval, CMMLU, and CCPM. 

Due to our efficient architectures and comprehensive engineering optimizations, \dsviii{} achieves extremely high training efficiency. 
Under our training framework and infrastructures, training \dsviii{} on each trillion tokens requires only 180K H800 GPU hours, which is much cheaper than training 72B or 405B dense models. 

\begin{table}[h]
    \centering
    \footnotesize
    \setlength{\tabcolsep}{8pt}
    \begin{tabular}{@{}l c | c c | c c@{}}
    \toprule
    \multirow{2}{*}{\centering \textbf{Benchmark (Metric)}} & \multirow{2}{*}{\textbf{\# Shots}} & \textbf{Small MoE} & \textbf{Small MoE} & \textbf{Large MoE} & \textbf{Large MoE} \\
     & & \textbf{Baseline} & \textbf{w/ MTP} & \textbf{Baseline} & \textbf{w/ MTP} \\
    \midrule
    \# Activated Params {\tiny (Inference)} & - & 2.4B & 2.4B & 20.9B & 20.9B \\
    \# Total Params {\tiny (Inference)} & - & 15.7B & 15.7B & 228.7B & 228.7B \\
    \# Training Tokens & - & 1.33T & 1.33T & 540B & 540B \\
    \midrule
    Pile-test {\tiny (BPB)} & - & \textbf{0.729} & \textbf{0.729} & 0.658 & \textbf{0.657} \\
    BBH {\tiny (EM)} & 3-shot & 39.0 & \textbf{41.4} & 70.0 & \textbf{70.7} \\
    MMLU {\tiny (EM)} & 5-shot & 50.0 & \textbf{53.3} & \textbf{67.5} & 66.6 \\
    DROP {\tiny (F1)} & 1-shot & 39.2 & \textbf{41.3} & 68.5 & \textbf{70.6} \\
    TriviaQA {\tiny (EM)} & 5-shot & 56.9 & \textbf{57.7} & \textbf{67.0} & \textbf{67.3} \\
    NaturalQuestions {\tiny (EM)} & 5-shot & \textbf{22.7} & 22.3 & 27.2 & \textbf{28.5} \\
    HumanEval {\tiny (Pass@1)} & 0-shot & 20.7 & \textbf{26.8} & 44.5 & \textbf{53.7} \\
    MBPP {\tiny (Pass@1)} & 3-shot & 35.8 & \textbf{36.8} & 61.6 & \textbf{62.2} \\
    GSM8K {\tiny (EM)} & 8-shot & 25.4 & \textbf{31.4} & 72.3 & \textbf{74.0} \\
    MATH {\tiny (EM)} & 4-shot & 10.7 & \textbf{12.6} & 38.6 & \textbf{39.8} \\
    \bottomrule
    \end{tabular}
    \caption{
    Ablation results for the MTP strategy. 
    The MTP strategy consistently enhances the model performance on most of the evaluation benchmarks.
    }
    \label{tab:ablation_nextn}
\end{table}

\subsection{Discussion}

\subsubsection{Ablation Studies for Multi-Token Prediction}
\label{discussion:ablation_nextn}

In Table~\ref{tab:ablation_nextn}, we show the ablation results for the MTP strategy. 
To be specific, we validate the MTP strategy on top of two baseline models across different scales. 
At the small scale, we train a baseline MoE model comprising 15.7B total parameters on 1.33T tokens. 
At the large scale, we train a baseline MoE model comprising 228.7B total parameters on 540B tokens. 
On top of them, keeping the training data and the other architectures the same, we append a 1-depth MTP module onto them and train two models with the MTP strategy for comparison. 
Note that during inference, we directly discard the MTP module, so the inference costs of the compared models are exactly the same. 
From the table, we can observe that the MTP strategy consistently enhances the model performance on most of the evaluation benchmarks. 

\subsubsection{Ablation Studies for the Auxiliary-Loss-Free Balancing Strategy}
\label{discussion:ablation_noaux_tc}

In Table~\ref{tab:ablation_noaux_tc}, we show the ablation results for the auxiliary-loss-free balancing strategy.
We validate this strategy on top of two baseline models across different scales. 
At the small scale, we train a baseline MoE model comprising 15.7B total parameters on 1.33T tokens. 
At the large scale, we train a baseline MoE model comprising 228.7B total parameters on 578B tokens. 
Both of the baseline models purely use auxiliary losses to encourage load balance, and use the sigmoid gating function with top-K affinity normalization. 
Their hyper-parameters to control the strength of auxiliary losses are the same as \dsvii{}-Lite and \dsvii{}, respectively. 
On top of these two baseline models, keeping the training data and the other architectures the same, we remove all auxiliary losses and introduce the auxiliary-loss-free balancing strategy for comparison. 
From the table, we can observe that the auxiliary-loss-free strategy consistently achieves better model performance on most of the evaluation benchmarks. 

\begin{table}[t]
    \centering
    \footnotesize
    \setlength{\tabcolsep}{4pt}
    \begin{tabular}{@{}l c | c c | c c@{}}
    \toprule
    \multirow{2}{*}{\centering \textbf{Benchmark (Metric)}} & \multirow{2}{*}{\textbf{\# Shots}} & \textbf{Small MoE} & \textbf{Small MoE} & \textbf{Large MoE} & \textbf{Large MoE} \\
     & & \textbf{Aux-Loss-Based} & \textbf{Aux-Loss-Free} & \textbf{Aux-Loss-Based} & \textbf{Aux-Loss-Free} \\
    \midrule
    \# Activated Params & - & 2.4B & 2.4B & 20.9B & 20.9B \\
    \# Total Params & - & 15.7B & 15.7B & 228.7B & 228.7B \\
    \# Training Tokens & - & 1.33T & 1.33T & 578B & 578B \\
    \midrule
    Pile-test {\tiny (BPB)} & - & 0.727 & \textbf{0.724} & 0.656 & \textbf{0.652} \\
    BBH {\tiny (EM)} & 3-shot & 37.3 & \textbf{39.3} & 66.7 & \textbf{67.9} \\
    MMLU {\tiny (EM)} & 5-shot & 51.0 & \textbf{51.8} & \textbf{68.3} & 67.2 \\
    DROP {\tiny (F1)} & 1-shot & 38.1 & \textbf{39.0} & \textbf{67.1} & \textbf{67.1} \\
    TriviaQA {\tiny (EM)} & 5-shot & \textbf{58.3} & \textbf{58.5} & 66.7 & \textbf{67.7} \\
    NaturalQuestions {\tiny (EM)} & 5-shot & \textbf{23.2} & \textbf{23.4} & 27.1 & \textbf{28.1} \\
    HumanEval {\tiny (Pass@1)} & 0-shot & 22.0 & \textbf{22.6} & 40.2 & \textbf{46.3} \\
    MBPP {\tiny (Pass@1)} & 3-shot & \textbf{36.6} & 35.8 & 59.2 & \textbf{61.2} \\
    GSM8K {\tiny (EM)} & 8-shot & 27.1 & \textbf{29.6} & 70.7 & \textbf{74.5} \\
    MATH {\tiny (EM)} & 4-shot & \textbf{10.9} & \textbf{11.1} & 37.2 & \textbf{39.6} \\
    \bottomrule
    \end{tabular}
    \caption{
    Ablation results for the auxiliary-loss-free balancing strategy. 
    Compared with the purely auxiliary-loss-based method, the auxiliary-loss-free strategy consistently achieves better model performance on most of the evaluation benchmarks.
    }
    \label{tab:ablation_noaux_tc}
\end{table}

\subsubsection{Batch-Wise Load Balance VS. Sequence-Wise Load Balance}
\label{discussion:balance}

\begin{figure}[!t]
\centering
\includegraphics[width=0.99\linewidth]{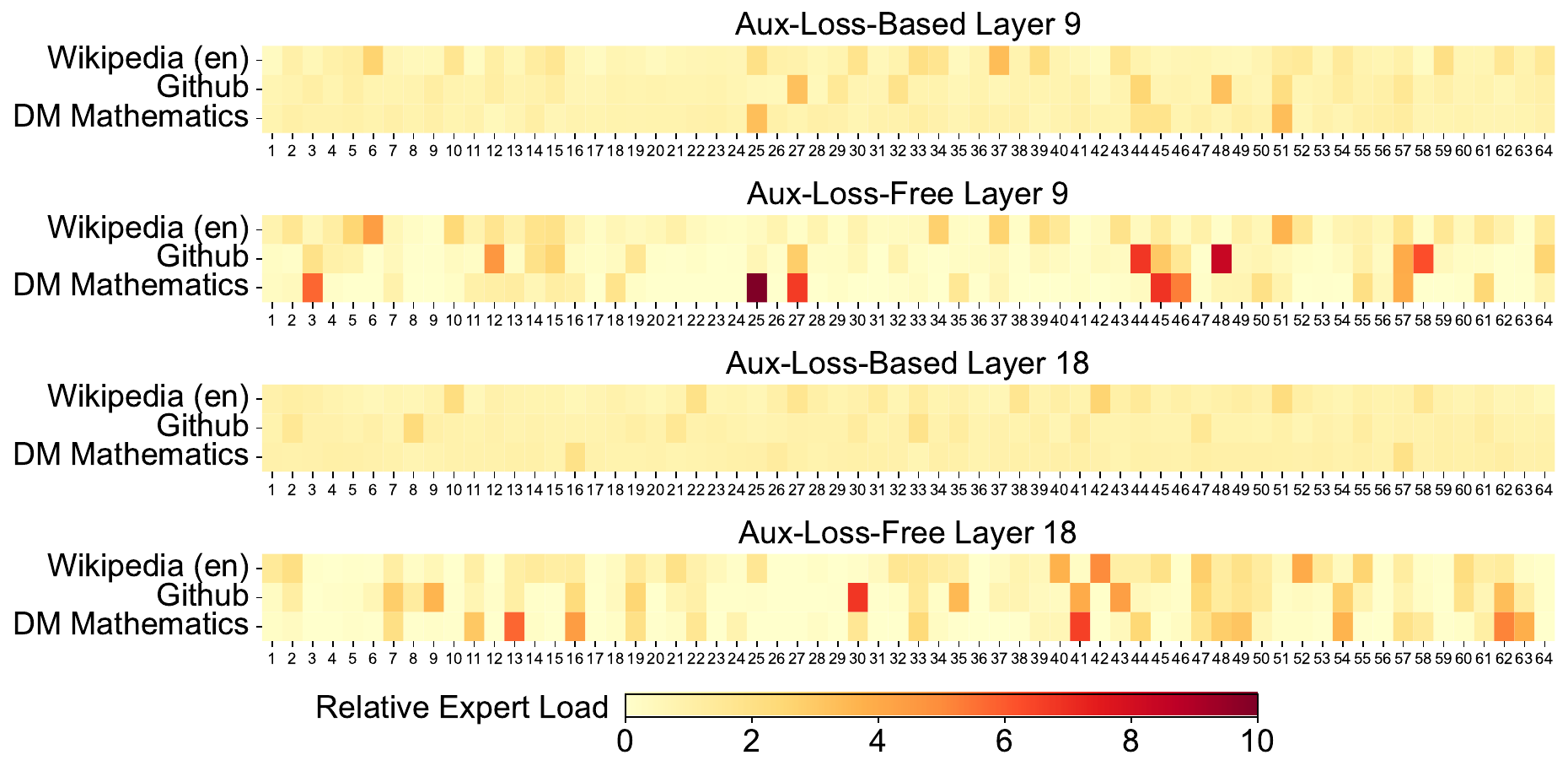}
\caption{
    Expert load of auxiliary-loss-free and auxiliary-loss-based models on three domains in the Pile test set. 
    The auxiliary-loss-free model shows greater expert specialization patterns than the auxiliary-loss-based one.
    The relative expert load denotes the ratio between the actual expert load and the theoretically balanced expert load. 
    Due to space constraints, we only present the results of two layers as an example, with the results of all layers provided in Appendix~\ref{app:detailed_expert_load}.
}
\label{fig:expert_load}
\end{figure}

The key distinction between auxiliary-loss-free balancing and sequence-wise auxiliary loss lies in their balancing scope: batch-wise versus sequence-wise. 
Compared with the sequence-wise auxiliary loss, batch-wise balancing imposes a more flexible constraint, as it does not enforce in-domain balance on each sequence. 
This flexibility allows experts to better specialize in different domains. 
To validate this, we record and analyze the expert load of a 16B auxiliary-loss-based baseline and a 16B auxiliary-loss-free model on different domains in the Pile test set.
As illustrated in Figure~\ref{fig:expert_load}, we observe that the auxiliary-loss-free model demonstrates greater expert specialization patterns as expected. 

To further investigate the correlation between this flexibility and the advantage in model performance, we additionally design and validate a batch-wise auxiliary loss that encourages load balance on each training batch instead of on each sequence. 
The experimental results show that, when achieving a similar level of batch-wise load balance, the batch-wise auxiliary loss can also achieve similar model performance to the auxiliary-loss-free method.
To be specific, in our experiments with 1B MoE models, the validation losses are: 2.258 (using a sequence-wise auxiliary loss), 2.253 (using the auxiliary-loss-free method), and 2.253 (using a batch-wise auxiliary loss). 
We also observe similar results on 3B MoE models: the model using a sequence-wise auxiliary loss achieves a validation loss of 2.085, and the models using the auxiliary-loss-free method or a batch-wise auxiliary loss achieve the same validation loss of 2.080.

In addition, although the batch-wise load balancing methods show consistent performance advantages, they also face two potential challenges in efficiency: 
(1) load imbalance within certain sequences or small batches, and 
(2) domain-shift-induced load imbalance during inference. 
The first challenge is naturally addressed by our training framework that uses large-scale expert parallelism and data parallelism, which guarantees a large size of each micro-batch. 
For the second challenge, we also design and implement an efficient inference framework with redundant expert deployment, as described in Section \ref{sec:inference_deployment}, to overcome it.

\section{Post-Training}
\label{sec:alignment}

\subsection{Supervised Fine-Tuning}

We curate our instruction-tuning datasets to include 1.5M instances spanning multiple domains, with each domain employing distinct data creation methods tailored to its specific requirements.

\paragraph{Reasoning Data.} 
For reasoning-related datasets, including those focused on mathematics, code competition problems, and logic puzzles, we generate the data by leveraging an internal DeepSeek-R1 model. 
Specifically, while the R1-generated data demonstrates strong accuracy, it suffers from issues such as overthinking, poor formatting, and excessive length. 
Our objective is to balance the high accuracy of R1-generated reasoning data and the clarity and conciseness of regularly formatted reasoning data.

To establish our methodology, we begin by developing an expert model tailored to a specific domain, such as code, mathematics, or general reasoning, using a combined Supervised Fine-Tuning (SFT) and Reinforcement Learning (RL) training pipeline. 
This expert model serves as a data generator for the final model. 
The training process involves generating two distinct types of SFT samples for each instance: the first couples the problem with its original response in the format of <problem, original response>, while the second incorporates a system prompt alongside the problem and the R1 response in the format of <system prompt, problem, R1 response>.

The system prompt is meticulously designed to include instructions that guide the model toward producing responses enriched with mechanisms for reflection and verification. 
During the RL phase, the model leverages high-temperature sampling to generate responses that integrate patterns from both the R1-generated and original data, even in the absence of explicit system prompts. 
After hundreds of RL steps, the intermediate RL model learns to incorporate R1 patterns, thereby enhancing overall performance strategically.

Upon completing the RL training phase, we implement rejection sampling to curate high-quality SFT data for the final model, where the expert models are used as data generation sources. 
This method ensures that the final training data retains the strengths of DeepSeek-R1 while producing responses that are concise and effective.

\paragraph{Non-Reasoning Data.} 
For non-reasoning data, such as creative writing, role-play, and simple question answering, we utilize DeepSeek-V2.5 to generate responses and enlist human annotators to verify the accuracy and correctness of the data.

\paragraph{SFT Settings.} 
We fine-tune \dsviii{}-Base for two epochs using the SFT dataset, using the cosine decay learning rate scheduling that starts at $5 \times 10^{-6}$ and gradually decreases to $1 \times 10^{-6}$. 
During training, each single sequence is packed from multiple samples. 
However, we adopt a sample masking strategy to ensure that these examples remain isolated and mutually invisible.

\subsection{Reinforcement Learning}

\subsubsection{Reward Model}

We employ a rule-based Reward Model (RM) and a model-based RM in our RL process.

\paragraph{Rule-Based RM.} 
For questions that can be validated using specific rules, we adopt a rule-based reward system to determine the feedback. 
For instance, certain math problems have deterministic results, and we require the model to provide the final answer within a designated format (e.g., in a box), allowing us to apply rules to verify the correctness. 
Similarly, for LeetCode problems, we can utilize a compiler to generate feedback based on test cases. 
By leveraging rule-based validation wherever possible, we ensure a higher level of reliability, as this approach is resistant to manipulation or exploitation.

\paragraph{Model-Based RM.} 
For questions with free-form ground-truth answers, we rely on the reward model to determine whether the response matches the expected ground-truth. 
Conversely, for questions without a definitive ground-truth, such as those involving creative writing, the reward model is tasked with providing feedback based on the question and the corresponding answer as inputs. 
The reward model is trained from the \dsviii{} SFT checkpoints. 
To enhance its reliability, we construct preference data that not only provides the final reward but also includes the chain-of-thought leading to the reward. 
This approach helps mitigate the risk of reward hacking in specific tasks.

\subsubsection{Group Relative Policy Optimization}

Similar to \dsvii{}~\citep{dsvii}, we adopt Group Relative Policy Optimization~(GRPO)~\citep{deepseekmath}, which foregoes the critic model that is typically with the same size as the policy model, and estimates the baseline from group scores instead. 
Specifically, for each question $q$, GRPO samples a group of outputs $\{o_1, o_2, \cdots, o_G\}$ from the old policy model $\pi_{\theta_{old}}$ and then optimizes the policy model $\pi_{\theta}$ by maximizing the following objective:
\begin{equation}
\begin{split}
    \mathcal{J}_{GRPO}(\theta) &= \mathbb{E}{[q \sim P(Q), \{o_i\}_{i=1}^G \sim \pi_{\theta_{old}}(O|q)]}  \\
    & \frac{1}{G}\sum_{i=1}^G \left( \min \left( \frac{\pi_\theta(o_i |q)}{\pi_{\theta_{old}}(o_i |q)} A_i, \text{clip} \left( \frac{\pi_\theta(o_i |q)}{\pi_{\theta_{old}}(o_i |q)}, 1 - \epsilon, 1 + \epsilon \right)  A_i \right) - \beta \mathbb{D}_{KL}\left(\pi_{\theta} || \pi_{ref}\right)\right) ,
\end{split}
\label{eq:GRPO-obj}
\end{equation}
\begin{equation}
    \mathbb{D}_{KL}\left(\pi_{\theta} || \pi_{ref}\right) = \frac{\pi_{ref}(o_i|q)}{\pi_{\theta}(o_i|q)}- \log\frac{\pi_{ref}(o_i|q)}{\pi_{\theta}(o_i|q)} - 1,
\end{equation}
where $\epsilon$ and $\beta$ are hyper-parameters; 
$\pi_{ref}$ is the reference model; 
and $A_i$ is the advantage, derived from the rewards $\{r_1, r_2, \ldots, r_G\}$ corresponding to the outputs within each group:
\begin{equation}
    A_i = \frac{r_i - {\operatorname{mean}(\{r_1, r_2, \cdots, r_G\})}}{{\operatorname{std}(\{r_1, r_2, \cdots, r_G\})}}.
\end{equation}

We incorporate prompts from diverse domains, such as coding, math, writing, role-playing, and question answering, during the RL process. 
This approach not only aligns the model more closely with human preferences but also enhances performance on benchmarks, especially in scenarios where available SFT data are limited.

\subsection{Evaluations}

\subsubsection{Evaluation Settings}

\paragraph{Evaluation Benchmarks.}
Apart from the benchmark we used for base model testing, we further evaluate instructed models on IFEval~\citep{IFeval}, FRAMES~\citep{frames}, LongBench v2~\citep{bai2024longbench2}, GPQA~\citep{gpqa}, SimpleQA~\citep{simpleqa}, C-SimpleQA~\citep{csimpleqa}, SWE-Bench Verified~\citep{swe_verified}, Aider~\footnote{\url{https://aider.chat}}, LiveCodeBench~\citep{livecodebench} (questions from August 2024 to November 2024), Codeforces~\footnote{\url{https://codeforces.com}}, Chinese National High School Mathematics Olympiad (CNMO 2024)\footnote{\url{https://www.cms.org.cn/Home/comp/comp/cid/12.html}}, and American Invitational Mathematics Examination 2024 (AIME 2024)~\citep{AIME2024}. 

\paragraph{Compared Baselines.}
We conduct comprehensive evaluations of our chat model against several strong baselines, including \dsvii{}-0506, DeepSeek-V2.5-0905, Qwen2.5 72B Instruct, LLaMA-3.1 405B Instruct, Claude-Sonnet-3.5-1022, and GPT-4o-0513. 
For the \dsvii{} model series, we select the most representative variants for comparison. 
For closed-source models, evaluations are performed through their respective APIs.

\paragraph{Detailed Evaluation Configurations.}
For standard benchmarks including MMLU, DROP, GPQA, and SimpleQA, we adopt the evaluation prompts from the simple-evals framework\footnote{\url{https://github.com/openai/simple-evals}}. 
We utilize the Zero-Eval prompt format~\citep{Lin_ZeroEval_A_Unified_2024} for MMLU-Redux in a zero-shot setting. 
For other datasets, we follow their original evaluation protocols with default prompts as provided by the dataset creators.
For code and math benchmarks, the HumanEval-Mul dataset includes 8 mainstream programming languages (Python, Java, Cpp, C\#, JavaScript, TypeScript, PHP, and Bash) in total. 
We use CoT and non-CoT methods to evaluate model performance on LiveCodeBench, where the data are collected from August 2024 to November 2024. 
The Codeforces dataset is measured using the percentage of competitors. 
SWE-Bench verified is evaluated using the agentless framework~\citep{agentless}. 
We use the ``diff'' format to evaluate the Aider-related benchmarks. 
For mathematical assessments, AIME and CNMO 2024 are evaluated with a temperature of 0.7, and the results are averaged over 16 runs, while MATH-500 employs greedy decoding. 
We allow all models to output a maximum of 8192 tokens for each benchmark. 

\begin{table}[h]
    \centering
    \footnotesize
    \setlength{\tabcolsep}{1.9pt}
    \begin{tabular}{@{}c l | c  c | c c c c| c@{}}
    \toprule
    & \multirow{2}{*}{\centering \textbf{Benchmark {\tiny (Metric)}}} & {\textbf{DeepSeek}} & {\textbf{DeepSeek}} & \textbf{Qwen2.5} & \textbf{LLaMA-3.1}  & \textbf{Claude-3.5-}  & \textbf{GPT-4o}& \textbf{DeepSeek} \\
    & & \textbf{V2-0506}& \textbf{V2.5-0905} & \textbf{72B-Inst.} & \textbf{405B-Inst.} & \textbf{Sonnet-1022}  & \textbf{0513} & \textbf{V3} \\
    \midrule
    & Architecture &  MoE & MoE & Dense & Dense &-&- & MoE \\
    & \# Activated Params & 21B & 21B & 72B & 405B& -&-& 37B \\
    & \# Total Params &  236B & 236B & 72B & 405B &-&-& 671B \\
    \midrule
    \multirow{8}{*}{English}&   MMLU {\tiny (EM)}  & 78.2 & 80.6 & 85.3& \textbf{88.6} & \textbf{88.3}&87.2 & \textbf{88.5} \\
     & MMLU-Redux {\tiny (EM)} &77.9 & 80.3 &  85.6 &86.2& \textbf{88.9}& 88.0 & \textbf{89.1}\\
    & MMLU-Pro {\tiny (EM)} & 58.5 &  66.2 & 71.6 &73.3 & \textbf{78.0} & 72.6 & 75.9\\
    & DROP {\tiny (3-shot F1)} &83.0 & 87.8 & 76.7 & 88.7 & 88.3 & 83.7 & \textbf{91.6}\\
    & IF-Eval {\tiny (Prompt Strict)} &57.7 & 80.6 & 84.1 & 86.0 & \textbf{86.5} & 84.3 & 86.1\\
    & GPQA-Diamond {\tiny (Pass@1)} & 35.3 & 41.3& 49.0 & 51.1& \textbf{65.0} & 49.9 & 59.1\\
    & SimpleQA {\tiny (Correct)} & 9.0 & 10.2 & 9.1 & 17.1& 28.4 & \textbf{38.2}& 24.9\\
     & FRAMES {\tiny (Acc.)} & 66.9 & 65.4 & 69.8 & 70.0 & 72.5 & \textbf{80.5} & 73.3 \\
     & LongBench v2 {\tiny (Acc.)} & 31.6 & 35.4 & 39.4 & 36.1 & 41.0 & 48.1 &\textbf{48.7} \\
    \midrule
    \multirow{5}{*}{Code} & HumanEval-Mul {\tiny (Pass@1)} & 69.3 &77.4 & 77.3 & 77.2 & {81.7} &80.5&\textbf{82.6}\\
    & LiveCodeBench {\tiny (Pass@1-COT)} &18.8 & 29.2 & 31.1 & 28.4 & 36.3& 33.4& \textbf{40.5} \\
    & LiveCodeBench {\tiny (Pass@1)} &20.3 & 28.4 & 28.7 & 30.1 & 32.8& 34.2& \textbf{37.6} \\
    & Codeforces {\tiny (Percentile)} & 17.5 & 35.6 & 24.8 & 25.3 & 20.3 & 23.6 & \textbf{51.6} \\
    & SWE Verified {\tiny (Resolved)} &-&22.6& 23.8 & 24.5 & \textbf{50.8}&38.8&42.0\\
    & Aider-Edit {\tiny (Acc.)} & 60.3& 71.6 & 65.4 & 63.9 & \textbf{84.2} &72.9&79.7 \\
    & Aider-Polyglot {\tiny (Acc.)}  & -& 18.2 & 7.6 & 5.8 & 45.3&16.0&\textbf{49.6} \\
    \midrule
    \multirow{3}{*}{Math} & AIME 2024 {\tiny (Pass@1)} & 4.6 & 16.7 & 23.3 & 23.3 & 16.0 & 9.3 & \textbf{39.2} \\
    & MATH-500 {\tiny (EM)} &  56.3 & 74.7 & 80.0 & 73.8 & 78.3 & 74.6&\textbf{90.2} \\
    & CNMO 2024 {\tiny (Pass@1)} & 2.8 & 10.8 & 15.9& 6.8& 13.1 & 10.8 &\textbf{43.2} \\
    \midrule
    \multirow{3}{*}{Chinese} & CLUEWSC {\tiny (EM)} & 89.9& 90.4 & \textbf{91.4} & 84.7 & 85.4 & 87.9 & 90.9\\
    & C-Eval {\tiny (EM)} & 78.6& 79.5 & 86.1 & 61.5 & 76.7 & 76.0 & \textbf{86.5}\\
     & C-SimpleQA {\tiny (Correct)}  & 48.5& 54.1 & 48.4 & 50.4 & 51.3 & 59.3 & \textbf{64.8}\\
    \bottomrule
    \end{tabular}
    \caption{
        Comparison between \dsviii{} and other representative chat models. 
        All models are evaluated in a configuration that limits the output length to 8K.
        Benchmarks containing fewer than 1000 samples are tested multiple times using varying temperature settings to derive robust final results.
        \dsviii{} stands as the best-performing open-source model, and also exhibits competitive performance against frontier closed-source models. 
    }
    \label{tab:chat}
\end{table}

\subsubsection{Standard Evaluation}

Table~\ref{tab:chat} presents the evaluation results, showcasing that \dsviii{} stands as the best-performing open-source model. 
Additionally, it is competitive against frontier closed-source models like GPT-4o and Claude-3.5-Sonnet. 

\paragraph{English Benchmarks.}
MMLU is a widely recognized benchmark designed to assess the performance of large language models, across diverse knowledge domains and tasks. 
\dsviii{} demonstrates competitive performance, standing on par with top-tier models such as LLaMA-3.1-405B, GPT-4o, and Claude-Sonnet 3.5, while significantly outperforming Qwen2.5 72B. 
Moreover, \dsviii{} excels in MMLU-Pro, a more challenging educational knowledge benchmark, where it closely trails Claude-Sonnet 3.5. 
On MMLU-Redux, a refined version of MMLU with corrected labels, \dsviii{} surpasses its peers. 
In addition, on GPQA-Diamond, a PhD-level evaluation testbed, \dsviii{} achieves remarkable results, ranking just behind Claude 3.5 Sonnet and outperforming all other competitors by a substantial margin.

In long-context understanding benchmarks such as DROP, LongBench v2, and FRAMES, \dsviii{} continues to demonstrate its position as a top-tier model. 
It achieves an impressive 91.6 F1 score in the 3-shot setting on DROP, outperforming all other models in this category. 
On FRAMES, a benchmark requiring question-answering over 100k token contexts, \dsviii{} closely trails GPT-4o while outperforming all other models by a significant margin. 
This demonstrates the strong capability of \dsviii{} in handling extremely long-context tasks. 
The long-context capability of \dsviii{} is further validated by its best-in-class performance on LongBench v2, a dataset that was released just a few weeks before the launch of DeepSeek V3. 
On the factual knowledge benchmark, SimpleQA, \dsviii{} falls behind GPT-4o and Claude-Sonnet, primarily due to its design focus and resource allocation. 
\dsviii{} assigns more training tokens to learn Chinese knowledge, leading to exceptional performance on the C-SimpleQA. 
On the instruction-following benchmark, \dsviii{} significantly outperforms its predecessor, \dsvii{}-series, highlighting its improved ability to understand and adhere to user-defined format constraints.

\paragraph{Code and Math Benchmarks.}
Coding is a challenging and practical task for LLMs, encompassing engineering-focused tasks like SWE-Bench-Verified and Aider, as well as algorithmic tasks such as HumanEval and LiveCodeBench. 
In engineering tasks, \dsviii{} trails behind Claude-Sonnet-3.5-1022 but significantly outperforms open-source models. 
The open-source \dsviii{} is expected to foster advancements in coding-related engineering tasks. 
By providing access to its robust capabilities, \dsviii{} can drive innovation and improvement in areas such as software engineering and algorithm development, empowering developers and researchers to push the boundaries of what open-source models can achieve in coding tasks. 
In algorithmic tasks, \dsviii{} demonstrates superior performance, outperforming all baselines on benchmarks like HumanEval-Mul and LiveCodeBench. 
This success can be attributed to its advanced knowledge distillation technique, which effectively enhances its code generation and problem-solving capabilities in algorithm-focused tasks.

On math benchmarks, \dsviii{} demonstrates exceptional performance, significantly surpassing baselines and setting a new state-of-the-art for non-o1-like models. 
Specifically, on AIME, MATH-500, and CNMO 2024, \dsviii{} outperforms the second-best model, Qwen2.5 72B, by approximately 10\% in absolute scores, which is a substantial margin for such challenging benchmarks.
This remarkable capability highlights the effectiveness of the distillation technique from DeepSeek-R1, which has been proven highly beneficial for non-o1-like models.

\paragraph{Chinese Benchmarks.}
Qwen and DeepSeek are two representative model series with robust support for both Chinese and English. 
On the factual benchmark Chinese SimpleQA, \dsviii{} surpasses Qwen2.5-72B by 16.4 points, despite Qwen2.5 being trained on a larger corpus compromising 18T tokens, which are 20\% more than the 14.8T tokens that \dsviii{} is pre-trained on.

On C-Eval, a representative benchmark for Chinese educational knowledge evaluation, and CLUEWSC (Chinese Winograd Schema Challenge), \dsviii{} and Qwen2.5-72B exhibit similar performance levels, indicating that both models are well-optimized for challenging Chinese-language reasoning and educational tasks.

\begin{table}[t]
    \centering
    \begin{tabular}{c | c c}
    \toprule
    \textbf{Model} & \textbf{Arena-Hard} & \textbf{AlpacaEval 2.0} \\
    \midrule
    DeepSeek-V2.5-0905 & 76.2 & 50.5\\
    Qwen2.5-72B-Instruct & 81.2 & 49.1  \\
    LLaMA-3.1 405B & 69.3 & 40.5\\
    GPT-4o-0513 & 80.4 &  51.1 \\
    Claude-Sonnet-3.5-1022 & 85.2  & 52.0 \\
    DeepSeek-V3 & \textbf{85.5} & \textbf{70.0} \\
    \bottomrule
    \end{tabular}
    \caption{
    English open-ended conversation evaluations. 
    For AlpacaEval 2.0, we use the length-controlled win rate as the metric. 
    }
    \label{tab:open} 
\end{table}

\subsubsection{Open-Ended Evaluation}

In addition to standard benchmarks, we also evaluate our models on open-ended generation tasks using LLMs as judges, with the results shown in Table~\ref{tab:open}. 
Specifically, we adhere to the original configurations of AlpacaEval 2.0~\citep{alpaca2.0} and Arena-Hard~\citep{li2024crowdsourced}, which leverage GPT-4-Turbo-1106 as judges for pairwise comparisons. 
On Arena-Hard, \dsviii{} achieves an impressive win rate of over 86\% against the baseline GPT-4-0314, performing on par with top-tier models like Claude-Sonnet-3.5-1022. 
This underscores the robust capabilities of \dsviii{}, especially in dealing with complex prompts, including coding and debugging tasks. 
Furthermore, \dsviii{} achieves a groundbreaking milestone as the first open-source model to surpass 85\% on the Arena-Hard benchmark. 
This achievement significantly bridges the performance gap between open-source and closed-source models, setting a new standard for what open-source models can accomplish in challenging domains.

Similarly, \dsviii{} showcases exceptional performance on AlpacaEval 2.0, outperforming both closed-source and open-source models. 
This demonstrates its outstanding proficiency in writing tasks and handling straightforward question-answering scenarios. 
Notably, it surpasses DeepSeek-V2.5-0905 by a significant margin of 20\%, highlighting substantial improvements in tackling simple tasks and showcasing the effectiveness of its advancements.

\subsubsection{\dsviii{} as a Generative Reward Model}

We compare the judgment ability of \dsviii{} with state-of-the-art models, namely GPT-4o and Claude-3.5. 
Table~\ref{tab:rewardbench} presents the performance of these models in RewardBench \citep{lambert2024rewardbench}.
\dsviii{} achieves performance on par with the best versions of GPT-4o-0806 and Claude-3.5-Sonnet-1022, while surpassing other versions. 
Additionally, the judgment ability of \dsviii{} can also be enhanced by the voting technique. 
Therefore, we employ \dsviii{} along with voting to offer self-feedback on open-ended questions, thereby improving the effectiveness and robustness of the alignment process.

\begin{table}[t]
    \centering
    \begin{tabular}{c | c c c c c}
    \toprule
    \textbf{Model} & \textbf{Chat} & \textbf{Chat-Hard} & \textbf{Safety} & \textbf{Reasoning} & \textbf{Average} \\
    \midrule
    GPT-4o-0513 & 96.6 & 70.4 & 86.7 & 84.9 & 84.7\\
    GPT-4o-0806 & 96.1 & 76.1 & 88.1 & 86.6 & 86.7\\
    GPT-4o-1120 & 95.8 & 71.3 & 86.2 & 85.2 & 84.6\\
    \midrule 
    Claude-3.5-sonnet-0620 & 96.4 & 74.0 & 81.6 & 84.7 & 84.2\\
    Claude-3.5-sonnet-1022 & 96.4 & 79.7 & 91.1 & 87.6 & 88.7\\
    \midrule 
    \dsviii{} & 96.9 & 79.8 & 87.0 & 84.3 & 87.0\\
    \dsviii{} (maj@6) & 96.9 & 82.6 & 89.5 & 89.2 & 89.6\\
    \bottomrule
    \end{tabular}
    \caption{Performances of GPT-4o, Claude-3.5-sonnet and \dsviii{} on RewardBench.}
    \label{tab:rewardbench} 
\end{table}

\subsection{Discussion}

\subsubsection{Distillation from DeepSeek-R1} 

We ablate the contribution of distillation from DeepSeek-R1 based on DeepSeek-V2.5. 
The baseline is trained on short CoT data, whereas its competitor uses data generated by the expert checkpoints described above. 

Table~\ref{tab:distill} demonstrates the effectiveness of the distillation data, showing significant improvements in both LiveCodeBench and MATH-500 benchmarks. 
Our experiments reveal an interesting trade-off: the distillation leads to better performance but also substantially increases the average response length. 
To maintain a balance between model accuracy and computational efficiency, we carefully selected optimal settings for \dsviii{} in distillation.

Our research suggests that knowledge distillation from reasoning models presents a promising direction for post-training optimization. 
While our current work focuses on distilling data from mathematics and coding domains, this approach shows potential for broader applications across various task domains. 
The effectiveness demonstrated in these specific areas indicates that long-CoT distillation could be valuable for enhancing model performance in other cognitive tasks requiring complex reasoning. 
Further exploration of this approach across different domains remains an important direction for future research.

\begin{table}[t]
    \centering
    \begin{tabular}{c|cc|cc}
    \toprule
    \multirow{2}{*}{\textbf{Model}} & \multicolumn{2}{c}{\textbf{LiveCodeBench-CoT}} & \multicolumn{2}{c}{\textbf{MATH-500}} \\

     & Pass@1 & Length & Pass@1 & Length\\
    \midrule
     DeepSeek-V2.5 Baseline & 31.1 & 718 & 74.6 & 769\\
     DeepSeek-V2.5 +R1 Distill & 37.4 & 783 &83.2& 1510\\
    \bottomrule
    \end{tabular}
    \caption{
    The contribution of distillation from DeepSeek-R1. 
    The evaluation settings of LiveCodeBench and MATH-500 are the same as in Table~\ref{tab:chat}.
    }
    \label{tab:distill} 
\end{table}

\subsubsection{Self-Rewarding}

Rewards play a pivotal role in RL, steering the optimization process. 
In domains where verification through external tools is straightforward, such as some coding or mathematics scenarios, RL demonstrates exceptional efficacy.
However, in more general scenarios, constructing a feedback mechanism through hard coding is impractical. 
During the development of \dsviii{}, for these broader contexts, we employ the constitutional AI approach~\citep{bai2022constitutional}, leveraging the voting evaluation results of \dsviii{} itself as a feedback source. 
This method has produced notable alignment effects, significantly enhancing the performance of \dsviii{} in subjective evaluations. 
By integrating additional constitutional inputs, \dsviii{} can optimize towards the constitutional direction. 
We believe that this paradigm, which combines supplementary information with LLMs as a feedback source, is of paramount importance. 
The LLM serves as a versatile processor capable of transforming unstructured information from diverse scenarios into rewards, ultimately facilitating the self-improvement of LLMs. 
Beyond self-rewarding, we are also dedicated to uncovering other general and scalable rewarding methods to consistently advance the model capabilities in general scenarios.

\subsubsection{Multi-Token Prediction Evaluation} 

Instead of predicting just the next single token, \dsviii{} predicts the next 2 tokens through the MTP technique. 
Combined with the framework of speculative decoding~\citep{speculative_google,speculative_xhm}, it can significantly accelerate the decoding speed of the model. 
A natural question arises concerning the acceptance rate of the additionally predicted token. 
Based on our evaluation, the acceptance rate of the second token prediction ranges between 85\% and 90\% across various generation topics, demonstrating consistent reliability. 
This high acceptance rate enables \dsviii{} to achieve a significantly improved decoding speed, delivering 1.8 times TPS (Tokens Per Second).

\section{Conclusion, Limitations, and Future Directions}
\label{sec:conclusion}

In this paper, we introduce \dsviii{}, a large MoE language model with 671B total parameters and 37B activated parameters, trained on 14.8T tokens. 
In addition to the \dsattn{} and \dsmoe{} architectures, it also pioneers an auxiliary-loss-free strategy for load balancing and sets a multi-token prediction training objective for stronger performance.
The training of \dsviii{} is cost-effective due to the support of FP8 training and meticulous engineering optimizations. 
The post-training also makes a success in distilling the reasoning capability from the DeepSeek-R1 series of models.
Comprehensive evaluations demonstrate that DeepSeek-V3 has emerged as the strongest open-source model currently available, and achieves performance comparable to leading closed-source models like GPT-4o and Claude-3.5-Sonnet. 
Despite its strong performance, it also maintains economical training costs. 
It requires only 2.788M H800 GPU hours for its full training, including pre-training, context length extension, and post-training. 

While acknowledging its strong performance and cost-effectiveness, we also recognize that \dsviii{} has some limitations, especially on the deployment.
Firstly, to ensure efficient inference, the recommended deployment unit for \dsviii{} is relatively large, which might pose a burden for small-sized teams.
Secondly, although our deployment strategy for \dsviii{} has achieved an end-to-end generation speed of more than two times that of \dsvii{}, there still remains potential for further enhancement.
Fortunately, these limitations are expected to be naturally addressed with the development of more advanced hardware.

DeepSeek consistently adheres to the route of open-source models with longtermism, aiming to steadily approach the ultimate goal of AGI (Artificial General Intelligence).
In the future, we plan to strategically invest in research across the following directions.
\begin{itemize}
    \item 
    We will consistently study and refine our model architectures, aiming to further improve both the training and inference efficiency, striving to approach efficient support for infinite context length. 
    Additionally, we will try to break through the architectural limitations of Transformer, thereby pushing the boundaries of its modeling capabilities.
    \item 
    We will continuously iterate on the quantity and quality of our training data, and explore the incorporation of additional training signal sources, aiming to drive data scaling across a more comprehensive range of dimensions.
    \item 
    We will consistently explore and iterate on the deep thinking capabilities of our models, aiming to enhance their intelligence and problem-solving abilities by expanding their reasoning length and depth.
    \item 
    We will explore more comprehensive and multi-dimensional model evaluation methods to prevent the tendency towards optimizing a fixed set of benchmarks during research, which may create a misleading impression of the model capabilities and affect our foundational assessment.
\end{itemize}

\bibliography{main}

\newpage
\appendix

\section*{Appendix}

\section{Contributions and Acknowledgments}

\definecolor{damaiblue}{RGB}{0, 0, 100}
\definecolor{damaigreen}{RGB}{0, 100, 0}
\definecolor{damaired}{RGB}{100, 0, 0}

\begin{multicols}{2} 
\noindent
\textbf{\color{damaiblue} Research \& Engineering} \\
\color{damaiblue} Aixin Liu \\
\color{damaiblue} Bing Xue \\
\color{damaiblue} Bingxuan Wang \\
\color{damaiblue} Bochao Wu \\
\color{damaiblue} Chengda Lu \\
\color{damaiblue} Chenggang Zhao \\
\color{damaiblue} Chengqi Deng \\
\color{damaiblue} Chenyu Zhang* \\
\color{damaiblue} Chong Ruan \\
\color{damaiblue} Damai Dai \\
\color{damaiblue} Daya Guo \\
\color{damaiblue} Dejian Yang \\
\color{damaiblue} Deli Chen \\
\color{damaiblue} Erhang Li \\
\color{damaiblue} Fangyun Lin \\
\color{damaiblue} Fucong Dai \\
\color{damaiblue} Fuli Luo* \\
\color{damaiblue} Guangbo Hao \\
\color{damaiblue} Guanting Chen \\
\color{damaiblue} Guowei Li \\
\color{damaiblue} H. Zhang \\
\color{damaiblue} Han Bao* \\
\color{damaiblue} Hanwei Xu \\
\color{damaiblue} Haocheng Wang* \\
\color{damaiblue} Haowei Zhang \\
\color{damaiblue} Honghui Ding \\
\color{damaiblue} Huajian Xin* \\
\color{damaiblue} Huazuo Gao \\
\color{damaiblue} Hui Qu \\
\color{damaiblue} Jianzhong Guo \\
\color{damaiblue} Jiashi Li \\
\color{damaiblue} Jiawei Wang* \\
\color{damaiblue} Jingchang Chen \\
\color{damaiblue} Jingyang Yuan \\
\color{damaiblue} Junjie Qiu \\
\color{damaiblue} Junlong Li \\
\color{damaiblue} Junxiao Song \\
\color{damaiblue} Kai Dong \\
\color{damaiblue} Kai Hu* \\
\color{damaiblue} Kaige Gao \\
\color{damaiblue} Kang Guan \\
\color{damaiblue} Kexin Huang \\
\color{damaiblue} Kuai Yu \\
\color{damaiblue} Lean Wang \\
\color{damaiblue} Lecong Zhang \\
\color{damaiblue} Liang Zhao \\
\color{damaiblue} Litong Wang \\
\color{damaiblue} Liyue Zhang \\
\color{damaiblue} Mingchuan Zhang \\
\color{damaiblue} Minghua Zhang \\
\color{damaiblue} Minghui Tang \\
\color{damaiblue} Panpan Huang \\
\color{damaiblue} Peiyi Wang \\
\color{damaiblue} Qiancheng Wang \\
\color{damaiblue} Qihao Zhu \\
\color{damaiblue} Qinyu Chen \\
\color{damaiblue} Qiushi Du \\
\color{damaiblue} Ruiqi Ge \\
\color{damaiblue} Ruisong Zhang \\
\color{damaiblue} Ruizhe Pan \\
\color{damaiblue} Runji Wang \\
\color{damaiblue} Runxin Xu \\
\color{damaiblue} Ruoyu Zhang \\
\color{damaiblue} Shanghao Lu \\
\color{damaiblue} Shangyan Zhou \\
\color{damaiblue} Shanhuang Chen \\
\color{damaiblue} Shengfeng Ye \\
\color{damaiblue} Shirong Ma \\
\color{damaiblue} Shiyu Wang \\
\color{damaiblue} Shuiping Yu \\
\color{damaiblue} Shunfeng Zhou \\
\color{damaiblue} Shuting Pan \\
\color{damaiblue} Tao Yun \\
\color{damaiblue} Tian Pei \\
\color{damaiblue} Wangding Zeng \\
\color{damaiblue} Wanjia Zhao* \\
\color{damaiblue} Wen Liu \\
\color{damaiblue} Wenfeng Liang \\
\color{damaiblue} Wenjun Gao \\
\color{damaiblue} Wenqin Yu \\
\color{damaiblue} Wentao Zhang \\
\color{damaiblue} Xiao Bi \\
\color{damaiblue} Xiaodong Liu \\
\color{damaiblue} Xiaohan Wang \\
\color{damaiblue} Xiaokang Chen \\
\color{damaiblue} Xiaokang Zhang \\
\color{damaiblue} Xiaotao Nie \\
\color{damaiblue} Xin Cheng \\
\color{damaiblue} Xin Liu \\
\color{damaiblue} Xin Xie \\
\color{damaiblue} Xingchao Liu \\
\color{damaiblue} Xingkai Yu \\
\color{damaiblue} Xinyu Yang \\
\color{damaiblue} Xinyuan Li \\
\color{damaiblue} Xuecheng Su \\
\color{damaiblue} Xuheng Lin \\
\color{damaiblue} Y.K. Li \\
\color{damaiblue} Y.Q. Wang \\
\color{damaiblue} Y.X. Wei \\
\color{damaiblue} Yang Zhang \\
\color{damaiblue} Yanhong Xu \\
\color{damaiblue} Yao Li \\
\color{damaiblue} Yao Zhao \\
\color{damaiblue} Yaofeng Sun \\
\color{damaiblue} Yaohui Wang \\
\color{damaiblue} Yi Yu \\
\color{damaiblue} Yichao Zhang \\
\color{damaiblue} Yifan Shi \\
\color{damaiblue} Yiliang Xiong \\
\color{damaiblue} Ying He \\
\color{damaiblue} Yishi Piao \\
\color{damaiblue} Yisong Wang \\
\color{damaiblue} Yixuan Tan \\
\color{damaiblue} Yiyang Ma* \\
\color{damaiblue} Yiyuan Liu \\
\color{damaiblue} Yongqiang Guo \\
\color{damaiblue} Yu Wu \\
\color{damaiblue} Yuan Ou \\
\color{damaiblue} Yuduan Wang \\
\color{damaiblue} Yue Gong \\
\color{damaiblue} Yuheng Zou \\
\color{damaiblue} Yujia He \\
\color{damaiblue} Yunfan Xiong \\
\color{damaiblue} Yuxiang Luo \\
\color{damaiblue} Yuxiang You \\
\color{damaiblue} Yuxuan Liu \\
\color{damaiblue} Yuyang Zhou \\
\color{damaiblue} Z.F. Wu \\
\color{damaiblue} Z.Z. Ren \\
\color{damaiblue} Zehui Ren \\
\color{damaiblue} Zhangli Sha \\
\color{damaiblue} Zhe Fu \\
\color{damaiblue} Zhean Xu \\
\color{damaiblue} Zhenda Xie \\
\color{damaiblue} Zhengyan Zhang \\
\color{damaiblue} Zhewen Hao \\
\color{damaiblue} Zhibin Gou \\
\color{damaiblue} Zhicheng Ma \\
\color{damaiblue} Zhigang Yan \\
\color{damaiblue} Zhihong Shao \\
\color{damaiblue} Zhiyu Wu \\
\color{damaiblue} Zhuoshu Li \\
\color{damaiblue} Zihui Gu \\
\color{damaiblue} Zijia Zhu \\
\color{damaiblue} Zijun Liu* \\
\color{damaiblue} Zilin Li \\
\color{damaiblue} Ziwei Xie \\
\color{damaiblue} Ziyang Song \\
\color{damaiblue} Ziyi Gao \\
\color{damaiblue} Zizheng Pan \\

\noindent
\textbf{\color{damaigreen} Data Annotation} \\
\color{damaigreen} Bei Feng \\
\color{damaigreen} Hui Li \\
\color{damaigreen} J.L. Cai \\
\color{damaigreen} Jiaqi Ni \\
\color{damaigreen} Lei Xu \\
\color{damaigreen} Meng Li \\
\color{damaigreen} Ning Tian \\
\color{damaigreen} R.J. Chen \\
\color{damaigreen} R.L. Jin \\
\color{damaigreen} Ruyi Chen \\
\color{damaigreen} S.S. Li \\
\color{damaigreen} Shuang Zhou \\
\color{damaigreen} Tianyu Sun \\
\color{damaigreen} X.Q. Li \\
\color{damaigreen} Xiangyue Jin \\
\color{damaigreen} Xiaojin Shen \\
\color{damaigreen} Xiaosha Chen \\
\color{damaigreen} Xiaowen Sun \\
\color{damaigreen} Xiaoxiang Wang \\
\color{damaigreen} Xinnan Song \\
\color{damaigreen} Xinyi Zhou \\
\color{damaigreen} Y.X. Zhu \\
\color{damaigreen} Yanhong Xu \\
\color{damaigreen} Yanping Huang \\
\color{damaigreen} Yaohui Li \\
\color{damaigreen} Yi Zheng \\
\color{damaigreen} Yuchen Zhu \\
\color{damaigreen} Yunxian Ma \\
\color{damaigreen} Zhen Huang \\
\color{damaigreen} Zhipeng Xu \\
\color{damaigreen} Zhongyu Zhang \\

\noindent
\textbf{\color{damaired} Business \& Compliance} \\
\color{damaired} Dongjie Ji \\
\color{damaired} Jian Liang \\
\color{damaired} Jin Chen \\
\color{damaired} Leyi Xia \\
\color{damaired} Miaojun Wang \\
\color{damaired} Mingming Li \\
\color{damaired} Peng Zhang \\
\color{damaired} Shaoqing Wu \\
\color{damaired} Shengfeng Ye \\
\color{damaired} T. Wang \\
\color{damaired} W.L. Xiao \\
\color{damaired} Wei An \\
\color{damaired} Xianzu Wang \\
\color{damaired} Xinxia Shan \\
\color{damaired} Ying Tang \\
\color{damaired} Yukun Zha \\
\color{damaired} Yuting Yan \\
\color{damaired} Zhen Zhang \\

\end{multicols} 

Within each role, authors are listed alphabetically by the first name. 
Names marked with * denote individuals who have departed from our team. 

\section{Ablation Studies for Low-Precision Training}
\label{app:fp8}

\begin{figure}[!h]
\centering
\includegraphics[width=0.95\linewidth]{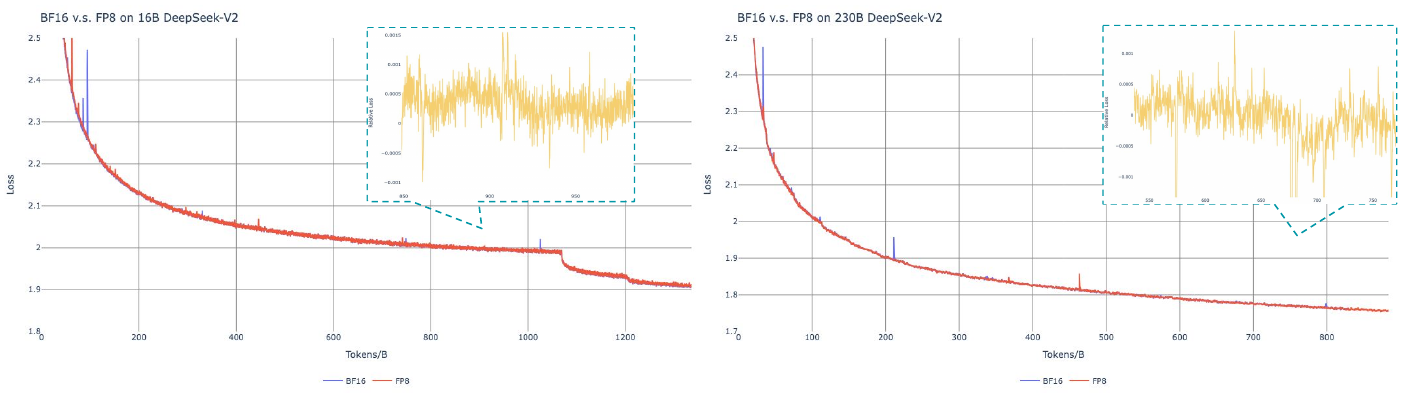}
\caption{
    Loss curves comparison between BF16 and FP8 training. 
    Results are smoothed by Exponential Moving Average (EMA) with a coefficient of 0.9.
}
\label{fig:fp8_vs_bf16}
\end{figure}

\subsection{FP8 v.s. BF16 Training}
\label{app:fp8_cp_bf16}
We validate our FP8 mixed precision framework with a comparison to BF16 training on top of two baseline models across different scales. At the small scale, we train a baseline MoE model comprising approximately 16B total parameters on 1.33T tokens. 
At the large scale, we train a baseline MoE model comprising approximately 230B total parameters on around 0.9T tokens. We show the training curves in Figure~\ref{fig:fp8_vs_bf16} and demonstrate that the relative error remains below 0.25\% with our high-precision accumulation and fine-grained quantization strategies.

\subsection{Discussion About Block-Wise Quantization}
\label{app:fp8_blockwise}
Although our tile-wise fine-grained quantization effectively mitigates the error introduced by feature outliers, it requires different groupings for activation quantization, i.e., \texttt{1x128} in forward pass and \texttt{128x1} for backward pass. 
A similar process is also required for the activation gradient. 
A straightforward strategy is to apply block-wise quantization per \texttt{128x128} elements like the way we quantize the model weights.
In this way, only transposition is required for backward.
Therefore, we conduct an experiment where all tensors associated with \texttt{Dgrad} are quantized on a block-wise basis. 
The results reveal that the \texttt{Dgrad} operation which computes the activation gradients and back-propagates to shallow layers in a chain-like manner, is highly sensitive to precision. 
Specifically, block-wise quantization of activation gradients leads to model divergence on an MoE model comprising approximately 16B total parameters, trained for around 300B tokens. 
We hypothesize that this sensitivity arises because activation gradients are highly imbalanced among tokens, resulting in token-correlated outliers~\citep{int4train}. 
These outliers cannot be effectively managed by a block-wise quantization approach.

\section{Expert Specialization Patterns of the 16B Aux-Loss-Based and Aux-Loss-Free Models}
\label{app:detailed_expert_load}
We record the expert load of the 16B auxiliary-loss-based baseline and the auxiliary-loss-free model on the Pile test set. 
The auxiliary-loss-free model tends to have greater expert specialization across all layers, as demonstrated in Figure~\ref{fig:detailed_expert_load}.

\begin{figure}[!t]
    \subfigure[Layers 1-7]{
        \includegraphics[width=0.95\linewidth]{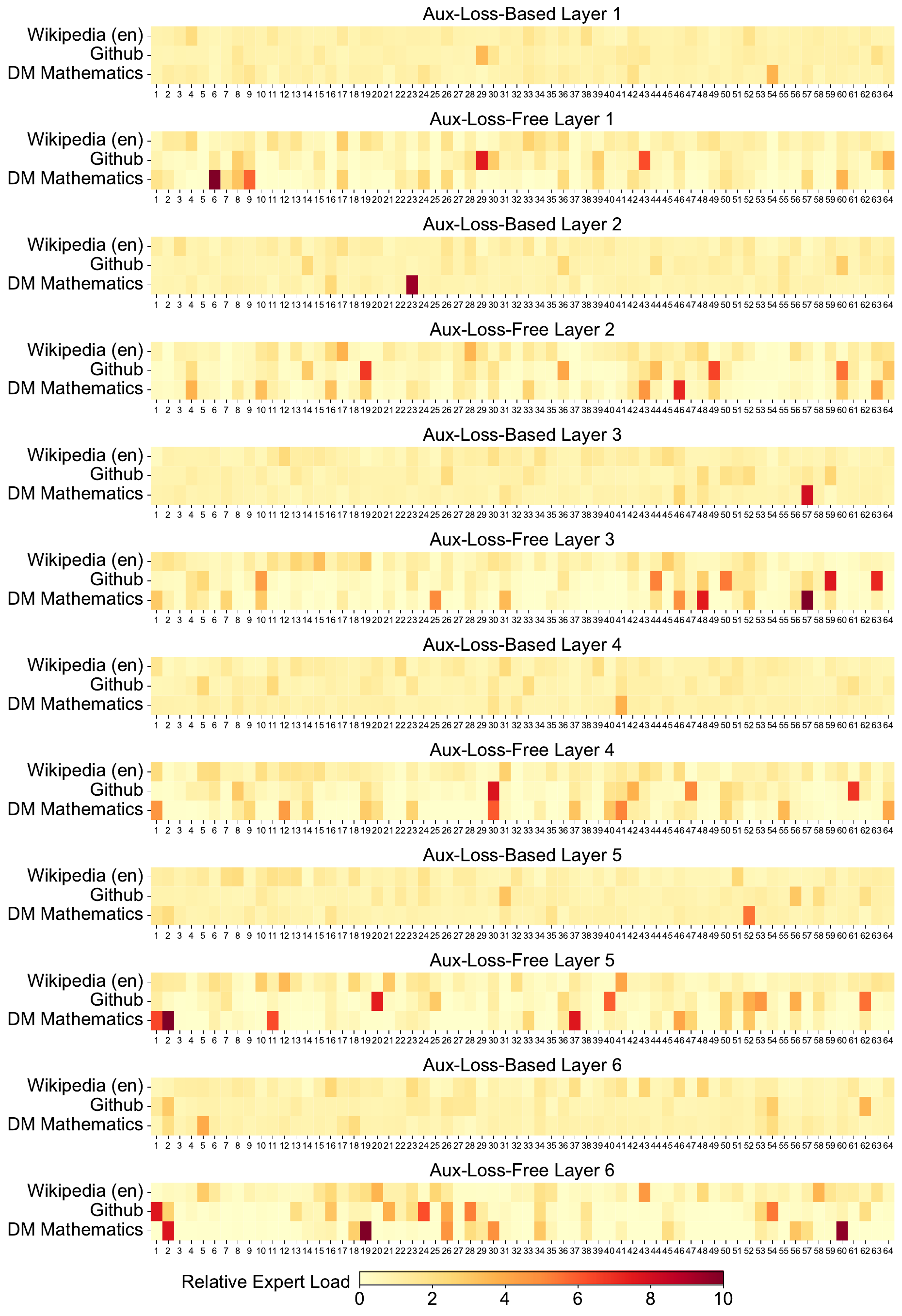}
    }
\end{figure}

\begin{figure}[!t]
    \ContinuedFloat
    \subfigure[Layers 7-13]{
        \includegraphics[width=0.95\linewidth]{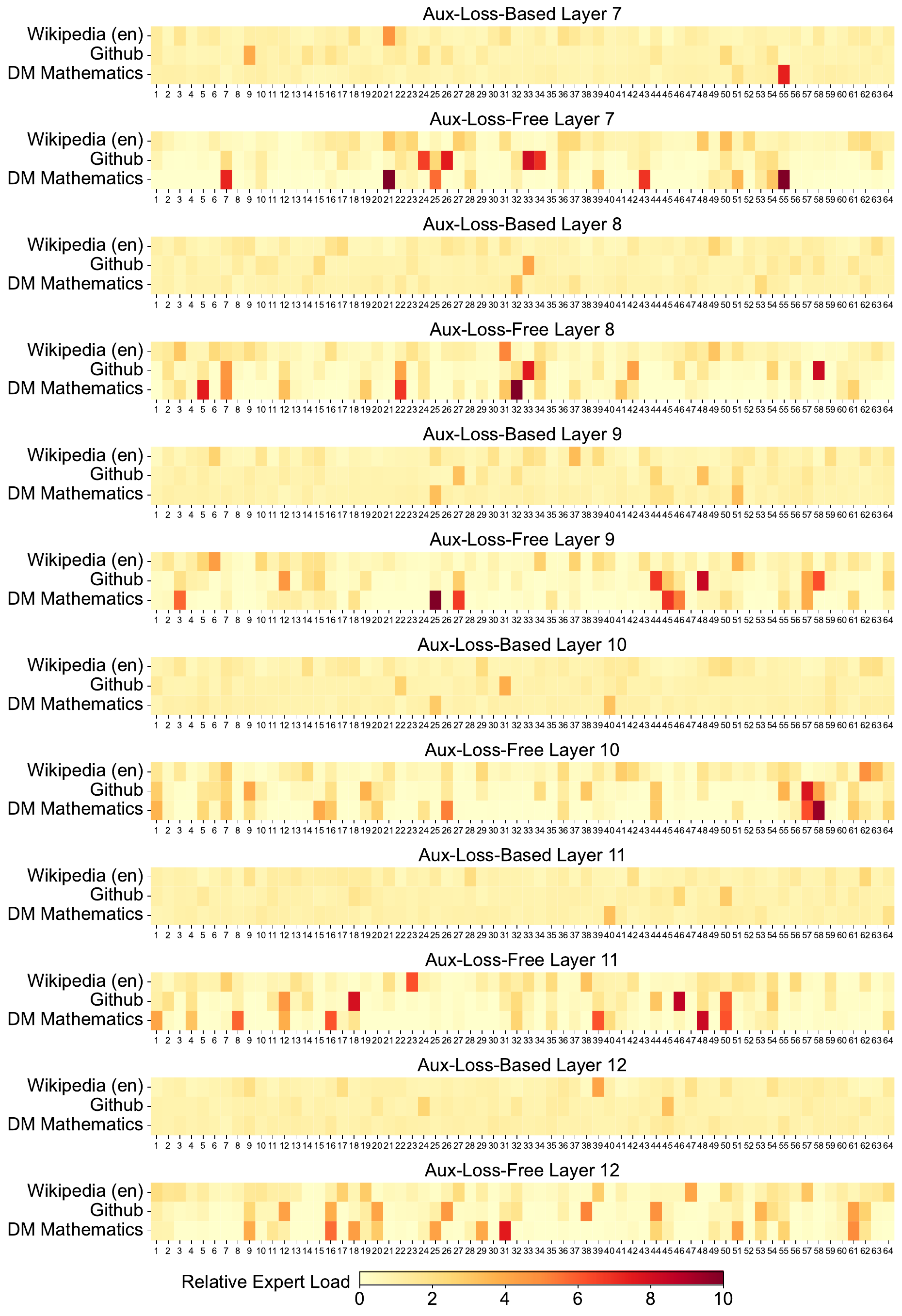}
    }
\end{figure}

\begin{figure}[!t]
    \ContinuedFloat
    \subfigure[Layers 13-19]{
        \includegraphics[width=0.95\linewidth]{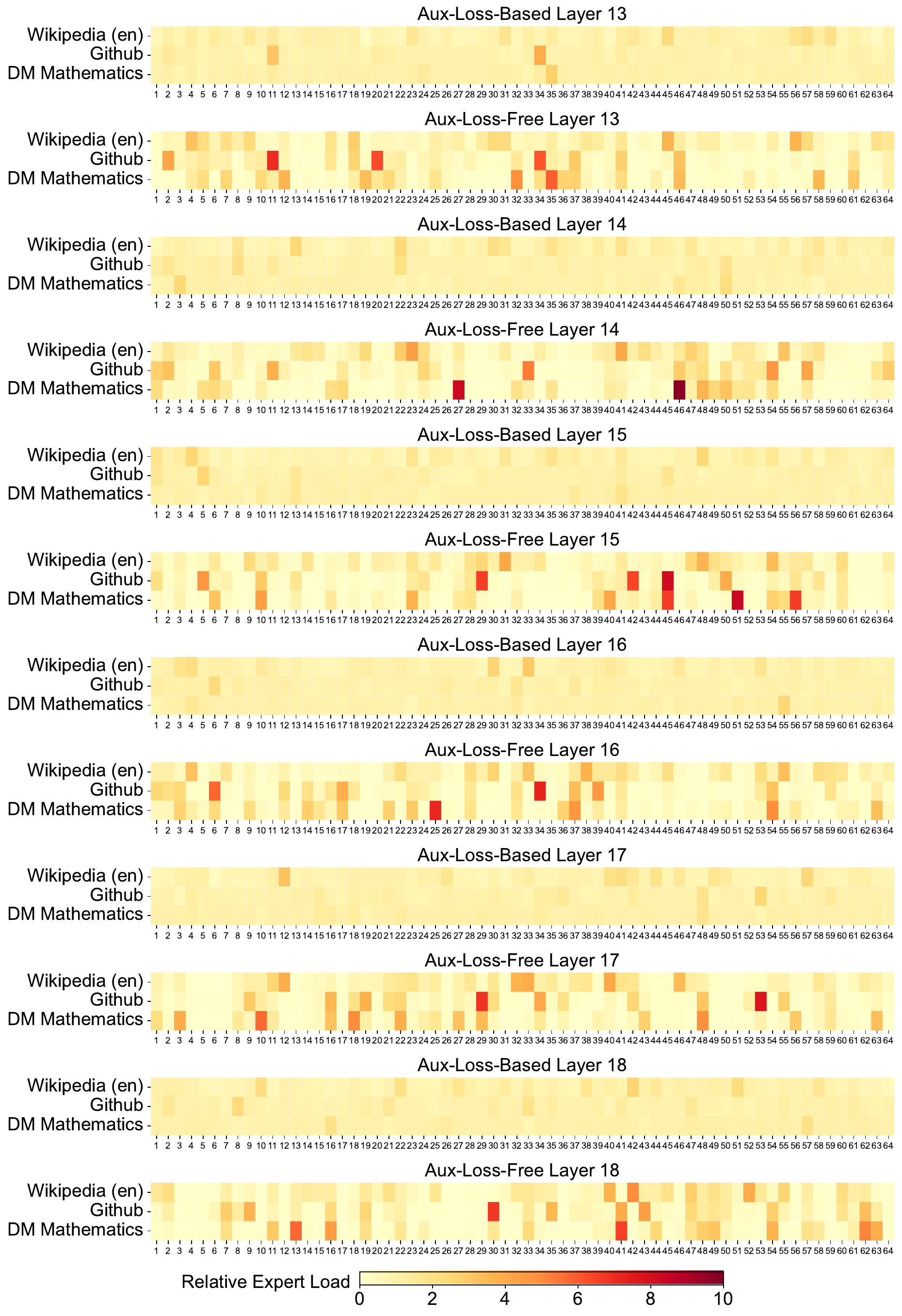}
    }
\end{figure}

\begin{figure}[!t]
    \ContinuedFloat
    \subfigure[Layers 19-25]{
        \includegraphics[width=0.95\linewidth]{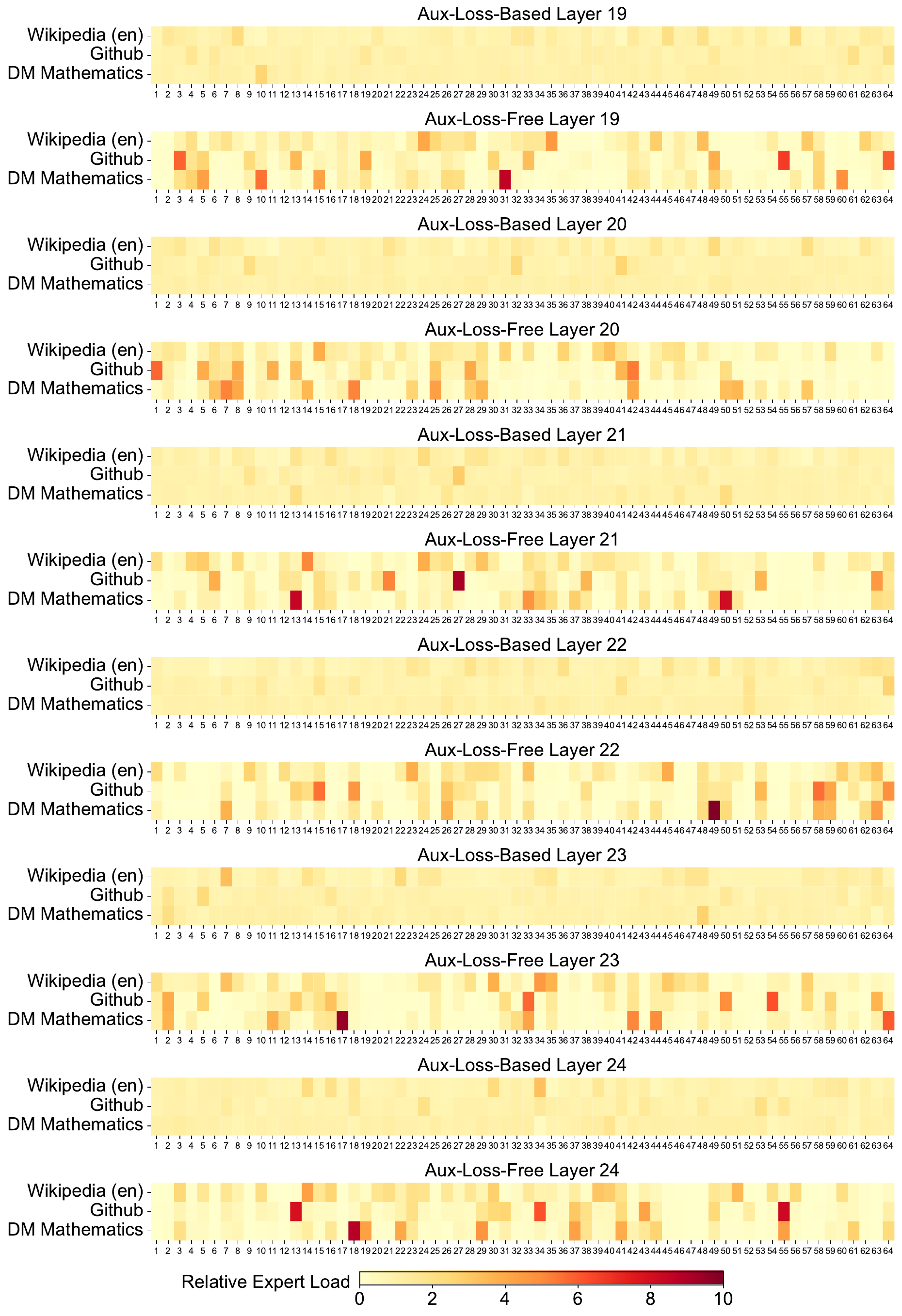}
    }
\end{figure}

\begin{figure}[!t]
    \ContinuedFloat
    \subfigure[Layers 25-27]{
        \includegraphics[width=0.95\linewidth]{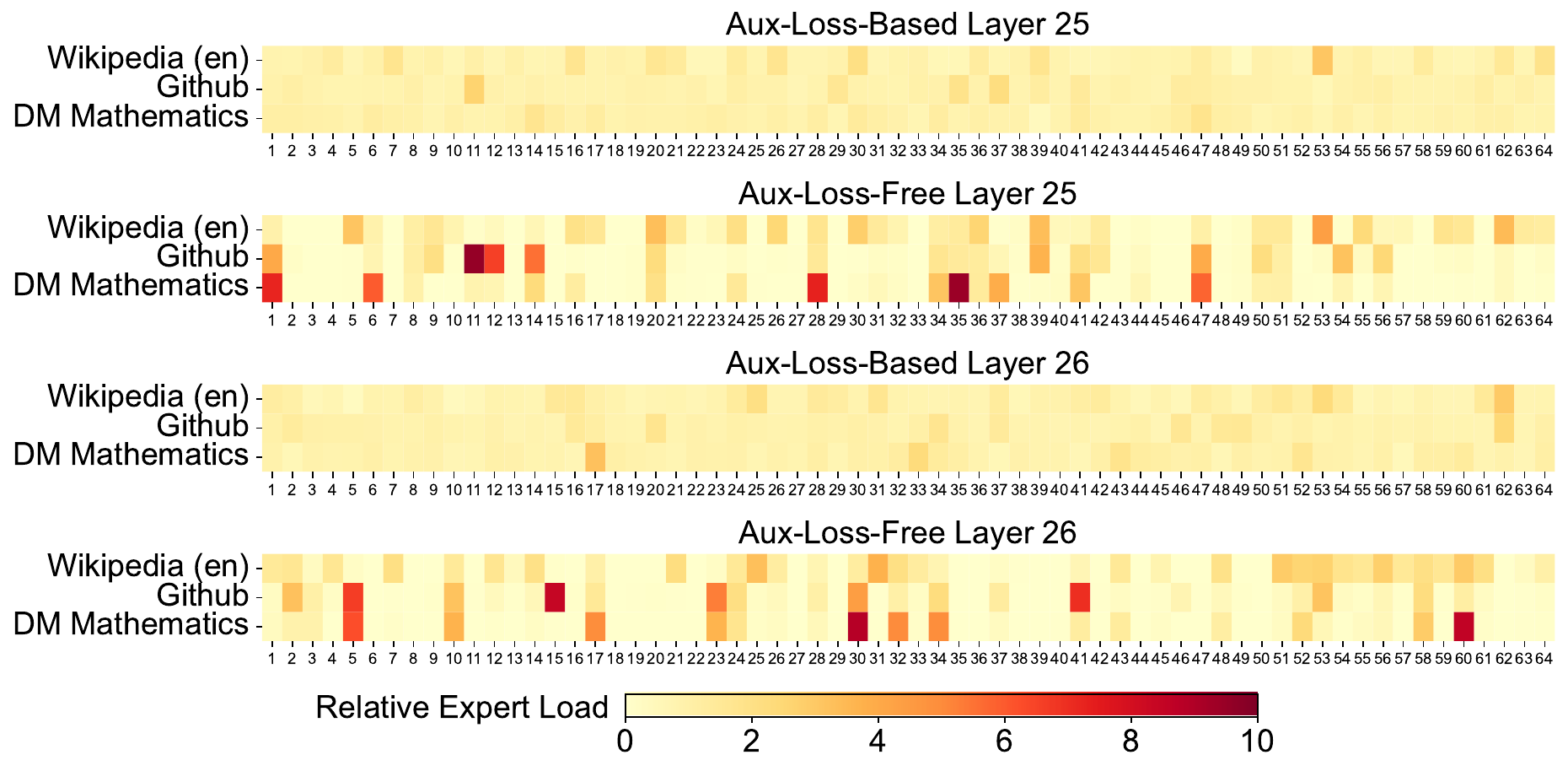}
    }
    \caption{
        Expert load of auxiliary-loss-free and auxiliary-loss-based models on three domains in the Pile test set. 
        The auxiliary-loss-free model shows greater expert specialization patterns than the auxiliary-loss-based one.
        The relative expert load denotes the ratio between the actual expert load and the theoretically balanced expert load. 
    }
\label{fig:detailed_expert_load}
\end{figure}

\newpage

\end{CJK*}
\end{document}